\documentclass[runningheads]{llncs}

\usepackage{amsmath}
\usepackage{amsfonts}
\usepackage{amssymb}
\usepackage{tabularx}
\usepackage{booktabs}
\usepackage{makecell}
\usepackage{svg}
\usepackage{graphicx}
\graphicspath{{./figures/}} %
\usepackage{csquotes}%
\usepackage{array}
\usepackage{todonotes}
\usepackage{algorithm}
\usepackage{algorithmicx}
\usepackage[noend]{algpseudocode}
\usepackage{enumitem}

\presetkeys{todonotes}{inline}{}
\usepackage{nicefrac}
\usepackage{multirow}
\usepackage{multicol}
\usepackage{array}  %
\usepackage{ragged2e}  %
\usepackage[sort&compress,capitalise,noabbrev]{cleveref}
\usepackage{cite} %
\usepackage{siunitx}
\usepackage[normalem]{ulem}
\robustify\bfseries
\robustify\uline

\newcommand{\ours}{MGMT}

\begin{document}
\title{Neural-Guided Equation Discovery}
\author{
Jannis Brugger\inst{1,2}\orcidID{0000-0002-7919-4789} \and
Mattia Cerrato\inst{4}\orcidID{0000-0001-7736-0547} \and
David Richter\inst{1}\orcidID{0009-0005-0649-4958} \and
Cedric Derstroff\inst{1,2}\orcidID{0000-0002-7475-7546} \and 
Daniel Maninger\inst{1,2}\orcidID{0009-0005-0649-4958} \and
Mira Mezini\inst{1,2,3}\orcidID{0000-0001-6563-7537} \and
Stefan Kramer\inst{4}\orcidID{0000-0003-0136-2540}
}
\authorrunning{J. Brugger et al.}
\institute{
  Technical University of Darmstadt, 64289 Darmstadt, Germany
\and
  Hessian Center for Artificial Intelligence (hessian.AI), 64293 Darmstadt, Germany
\and National Research Center for Applied Cybersecurity ATHENE
\and
  Johannes Gutenberg-Universität Mainz, 55128 Mainz, Germany
}

\maketitle              %

\begin{abstract}
Deep learning approaches are becoming increasingly attractive for equation discovery. We show the advantages and disadvantages of using neural-guided equation discovery by giving an overview of recent papers and the results of experiments using our modular equation discovery system \ours{} (\textbf{M}ulti-Task \textbf{G}rammar-Guided \textbf{M}onte-Carlo \textbf{T}ree Search for Equation Discovery). The system uses neural-guided Monte-Carlo Tree Search (MCTS) and supports both supervised and reinforcement learning, with a search space defined by a context-free grammar.
We summarize seven desirable properties of equation discovery systems, emphasizing the importance of embedding tabular data sets for such learning approaches. 
Using the modular structure of \ours{}, we compare seven architectures (among them, RNNs, CNNs, and Transformers) for embedding tabular datasets on the auxiliary task of contrastive learning for tabular data sets on an equation discovery task. For almost all combinations of modules, supervised learning outperforms reinforcement learning. Moreover, our experiments indicate an advantage of using grammar rules as action space instead of tokens. Two adaptations of MCTS -- risk-seeking MCTS and AmEx-MCTS -- can improve equation discovery with that kind of search.
\keywords{Equation Discovery \and Neural-Guided Search \and Context-Free Grammars \and Monte-Carlo Tree Search}
\end{abstract}

\setcounter{tocdepth}{1}
\clearpage{} 

\section{Introduction}

{\em Equation discovery (ED)} \cite{Langley77,langley_scientific_1987,DzeroskiTodorovski1993,TodorovskiDzeroski1997,Washio1999,Todorovski2005,Ganzert2010} and {\em Symbolic Regression (SR)} \cite{koza1994genetic,SchmidtLipson2009} have evolved from small research fields with pioneering work to fields with considerable traction and progress \cite{Probabilistic_grammars_for_equation_discovery,udrescu_ai_2020-1,Deep_symbolic_regression_Recovering_mathematical_expressions_from_data_via_risk_seeking_policy_gradients}.
Early approaches to equation discovery and symbolic regression had to limit  search or employed genetic programming \cite{koza1994genetic} to stochastically search very large search spaces, which can be time-consuming, with uncertain outcome. Recent advancements of neural network guided search \cite{silver2018alphazero} hint at possible improvements. In fact, a plethora of different approaches exist that employ neural networks to guide the search for suitable candidate equations (e.g., DSR \cite{Deep_symbolic_regression_Recovering_mathematical_expressions_from_data_via_risk_seeking_policy_gradients} or AI Feynman 2.0 \cite{udrescu_ai_2020-1}). It is unclear, however, how to best use neural networks in this effort (e.g., directly, via supervised learning or via Monte-Carlo Tree Search (MCTS)  \cite{coulom_efficient_2007,silver2018alphazero}). Also, it is unclear whether grammars are still useful in search in this context, or whether the prediction should be based on tokens, as is done in many systems today (see Table \ref{table_approaches}).

The purpose of this chapter is to study questions around the use of neural networks for guiding the search for candidate equations. Our vehicle for answering these questions is a system called {\ours{} (\textbf{M}ulti-Task \textbf{G}rammar-Guided \textbf{M}onte-Carlo \textbf{T}ree Search for Equation Discovery)}, which integrates elements of MCTS \cite{coulom_efficient_2007}, guided by a neural network (akin to AlphaZero \cite{silver2018alphazero}), and grammar-based equation discovery \cite{TodorovskiDzeroski1997,Probabilistic_grammars_for_equation_discovery}. Some of the below questions will also be partially answered by reference to existing literature.

\begin{figure}[t]
  \includegraphics[width=\linewidth]{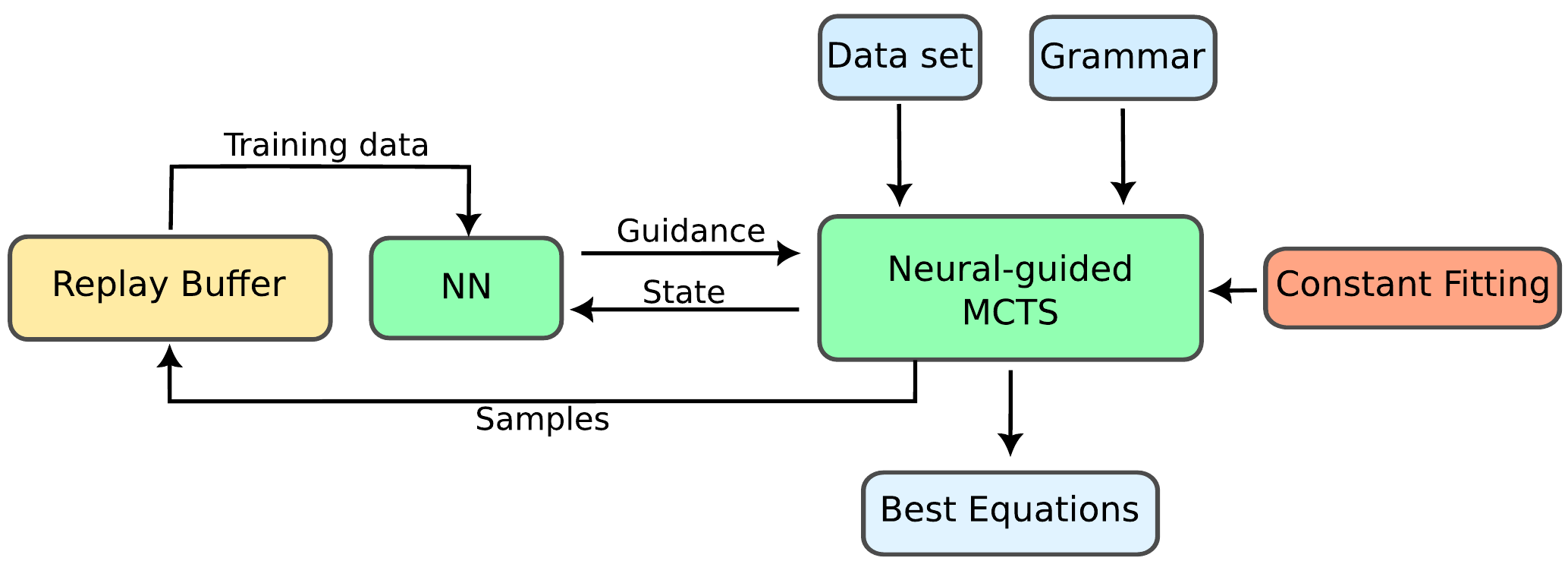}
  \caption{\textbf{Overview \ours{}} A grammar and a data set are the input in \ours{}. The search is performed by a neural-guided MCTS, which receives guidance from the neural network for each state in the search. Once a complete equation has been generated, all constants in the equation are fitted. The results of the MCTS are stored in a replay buffer and used to train the neural network. The aim is for the neural network and the MCTS to reinforce each other. The output of the system are the $k$-best equations found during the search.}
  \label{fig:overview_neural_guided_mcts}
\end{figure}

The research questions discussed as part of this chapter are: 
\begin{itemize}
    \item \textbf{R1} Can a search guided by neural networks provide results of the same quality with fewer visited states compared to a search without neural networks?
    \item \textbf{R2} Is an MCTS-based training better than a supervised training and can MCTS be adapted for the domain of equation discovery to be more efficient? 
        \item \textbf{R3} Can the integration of data set embeddings improve the search, and what is the best way of doing so?
    \item \textbf{R4} Is it advantageous to restrict the search space using grammars in neural-guided equation discovery, or should the search be based on tokens?

\end{itemize}
To answer these questions, we formulate the search for equations as a tree search as in a game, where the possible actions are given by a \emph{context-free grammar} \cite{CHOMSKY1959137} and the goal is to find the equation which fits the measurements as accurately as possible.
We use a neural guided Monte Carlo Tree search (MCTS) \cite{coulom_efficient_2007}, inspired by AlphaZero \cite{silver2018alphazero}, conducting the search. A high-level overview of our approach is given in \cref{fig:overview_neural_guided_mcts}. The guiding network obtains for each state in the search tree the current equation and the data set for which the symbolic description is sought. The guidance is a distribution regarding which grammar rule to apply next. Within this scheme, we will conduct experiments to shed some light on the above research questions.

The structure of this chapter is organized as follows: 
In \cref{sec:requirments_equation_discovery_system}, we formulate seven desirable properties of equation discovery systems. In \cref{sec:related_work}, we examine current equation discovery systems with respect to our proposed requirements. 

The details of \ours{} are explained in \cref{sec:methods}. In the same section, we suggest two adaptations of the ``pure'' MCTS, AmEX-MCTS and Risk-seeking MCTS. We also present contrastive learning for tabular data sets as an unsupervised auxiliary task to test the embeddings of NN architectures. In preparation for the experiments, \cref{sec:generation_of_training_data_sets} explains the generation of training sets using a context-free grammar.
In \cref{sec:experiments}, we evaluate our system experimentally and summarize the results with regard to the research questions. 

Limitations of the proposed methods and an outlook on relevant questions in the field of equation discovery are given in \cref{sec:limitations_and_outview}.
The chapter is concluded with a summary in \cref{sec:Conclusion}.

\section{Desirable Properties of Equation Discovery Systems} \label{sec:requirments_equation_discovery_system} 

From a scan of current literature (among others, articles cited in Table \ref{table_approaches}) and own considerations, we compiled a list of desirable properties of equation discovery systems. This list is neither meant to be comprehensive or complete, nor do we claim novelty. However, it will be useful for discussing some of the design decisions we took when developing the \ours{} system. 

The first four requirements relate to the search space. The search space for the modification of equations to better fit the data is vast: Considering an expression modeled as a tree of operators and variables/constants, the number of ways a tree can be configured grows exponentially with the depth of the tree. Independently of the goal of the current search, with every search we can learn something about the structure of the search space. To avoid having to re-explore the search space for each new problem from scratch, an equation discovery system should \textbf{(i) be able to take advantage of results from previous searches}. The notion of an AI system improving from task to task is as old as 1957 (as mentioned in the second letter from Herbert Simon to Bertrand Russell \cite{Simon1996}), and is currently popular under the heading of continual learning \cite{Wang2024}.

To achieve the previous point, the system has to make decisions with a \textbf{(ii) dependency on the data set}, because it is the only resource available before the search begins to estimate the similarity between two problems and to develop a heuristic where to start with the exploration. This heuristic should be updated during the search, e.g., by a score of how good the proposed equation so far fits the target data set.

Many equations are syntactically correct, however, are not of interest to the user, as they include parts that do not conform with prior knowledge.
By making \textbf{(iii) domain knowledge accessible} to the system, the search space is reduced and thus a solution useful to the user can be found faster. Well-known, well-established approaches to this are the use of grammars \cite{DzeroskiTodorovski1993,TodorovskiDzeroski1997} or libraries of process models \cite{Todorovski2005}.

\textbf{(iv) The search algorithm should be resource-efficient and should be able to operate under bounded resources.} In other words, it should be able to explore the search space with the available resources efficiently. 

The next three requirements are concerned with the relationship between the equation discovery system, the user, and the real world. 
The user should be able to \textbf{(v) trade off error and complexity} of a solution. Error refers to the differences between actual and predicted values.  
Complexity refers to the number of components of an equation. 
If an equation discovery system \textbf{(vi) is interactive}, the scientist and the system can mutually support each other, e.g., by the scientist indicating which parts of the equation are meaningful or the system presenting which data fit which models and where errors may have occurred. 
For a system that can help with real unsolved problems, \textbf{(vii) noise tolerance} is a necessary requirement. 
In the next chapter, we will examine how previous systems implement the first four requirements. 

\section{Related Work}\label{sec:related_work}

In the following, we will give an overview of recent work in equation discovery and point out the differences from \ours{}. 
Automated discovery systems differ in many dimensions: the basic learning technique, the representation of choice, and which parts of the scientific process they seek to automate or support. 
We refer the reader to recent surveys \cite{kramer2023automated,Makke2024} for an overview of these themes and differences between systems. 
For the purpose of this work, we pay particular attention to the problem of \emph{continual learning} \cite{Wang2024,chen2018lifelong}.
In the paradigm of continual learning, an intelligent system is evaluated for its capabilities to learn over time, possibly from a variety of different tasks.
Our interest here is to investigate whether it is possible for an automated discovery system to learn from its \emph{experience of discovery}. In other words, we seek to identify whether prior exposure (experience) to discovery problems is also beneficial to an equation discovery system.
In this section, we thus review existing approaches to equation discovery by dividing them into two broad categories: \emph{single-task} methods, which focus on one problem at a time and start from scratch every time, and \emph{continual, multi-task} methods, which seek to gather and transfer experience from one task to another. We will focus in particular on recent developments in the field.

\subsection{Single-Task Equation Discovery}

As previously mentioned, one group of methodologies in equation discovery seeks to learn equations from single data sets, without any transfer of knowledge from one task to another or any explicit way to gather experience from one task to another. 
At each step, the currently proposed equation is evaluated against a validation set; based on the obtained signal, it is then considered whether the change compared to the previously proposed equations was positive or negative and therefore further changes are made in the same direction or elsewhere. 
The techniques of reference for these methodologies are therefore Genetic Programming and, more generally, evolutionary algorithms. 
The general-purpose symbolic regression library \emph{PySR} \cite{cranmer2023interpretable} extends the mutation loop proposed by Koza \cite{koza1994genetic} by tying the selection probability of the fittest individual (equation) to the increase in fit quality via simulated annealing. 
Moreover, it includes subroutines for algebraic simplification of candidate equations and constant fitting.
In the same vein, \emph{Operon}
\cite{burlacu_operon_2020} introduces an efficient, extensible C++ implementation of genetic programming with particular attention given to thread-level parallelism. 
\emph{HVAE} \cite{Me_nar_2023} employs a hierarchical variational autoencoder to embed equation trees into a latent space, which is then explored via evolutionary techniques. 
The focus on modern connectionist systems is also present in \emph{DSR} \cite{Deep_symbolic_regression_Recovering_mathematical_expressions_from_data_via_risk_seeking_policy_gradients}, where a Recurrent Neural Network (RNN) is employed to output tokens that form an equation. The list of output tokens is then restructured into an equation syntax tree by the assumption that the list represents the pre-order traversal of a tree. 

The core innovation here is to employ what the authors call \emph{risk-seeking gradients}, that is, they reformulate the training objective of the recurrent network so that it may focus on the \emph{best performing} equations output by their model rather than computing some loss based on the average fitness of all equations. 
An extension of this approach by means of a genetic programming subroutine that randomly mutates candidate equations was proposed by Mundhenk \emph{et al.} \cite{mundhenk_symbolic_2021}, showing improved results.
Another line of investigation employs the Bayesian learning paradigm of posterior updating (\emph{Machine scientist} \cite{guimera_bayesian_2020} and $BSR$ \cite{jin_bayesian_2019}).
The main challenges in this space are choosing appropriate ways of subsuming distributions from the tree representation of equations and designing an update procedure.
Both the aforementioned methods employ Markov Chain Monte Carlo for the updates. The authors of the \emph{Machine scientist} \cite{guimera_bayesian_2020} propose three rules to modify a candidate equation tree, and take the integral over all possible equation parameters as the posterior distribution of one equation.
The authors also outline a procedure to learn priors from a corpus of equations.
\emph{BSR} \cite{jin_bayesian_2019} contrarily uses seven equation modification rules (six plus a ``start'' rule) and employ the Metropolis-Hastings algorithm to perform inference over a posterior distribution. 
The posterior distribution is taken to be the ordinary least squares fit of different candidate equations (trees), partially avoiding the problem of equation parameter learning. 

As Landajuela \emph{et al.} \cite{landajuela_unified_2022} have already established, orthogonal to the question if learning takes place across data sets or not, it can be considered whether an approach is model-free, i.e., works directly on the data, or is model-based, i.e., creates a model for the data. 

Examples of model-based approaches are \emph{AI Feynman} \cite{udrescu_ai_2020} and \emph{AI Feynman 2.0}\cite{udrescu_ai_2020-1}. In this line of work, a neural network is trained to approximate the function that generated the data set. 
This function is then checked for mathematical properties such as symmetry or separability to decompose the original equation discovery task into smaller and simpler tasks. Lusch \emph{et al.}
\cite{lusch_deep_2018} use an auto-encoder architecture to find nonlinear coordinates at which the dynamics of the system of interest are globally linear, in order to find a differential equation for these coordinates. 
\emph{EQL}$^{\div}$ \cite{sahoo_learning_2018} learns a shallow neural network architecture with specialized units for computation of primitive functions (e.g., trigonometric functions or other functions which are assumed to be relevant to the discovery problem). Therefore, in that approach, the network itself represents a \emph{model} of the discovery task and may be interpreted directly to obtain a symbolic equation.

\begin{table}[!pt]
\caption{\textbf{Overview of approaches using continual learning for equation discovery}. The abbreviation 3Token is used for approaches where each number is represented by 3 tokens: sign, mantissa $\left[0, 9999\right]$ and exponent $\left[E{-}100, E100\right]$}
\label{table_approaches}
\begin{tabularx}{\textwidth}{>{\centering\arraybackslash}X >{\centering\arraybackslash}X >{\centering\arraybackslash}X >{\centering\arraybackslash}X}
\toprule
   Paper  & Definition of search space & Representation data set & Algorithm  \\  
\midrule
NeSymReS \cite{biggio_neural_2021} & Next token prediction & Set Transformer \cite{lee_set_2019} & Sequence to sequence transformer  \\  
\hline
E2E \cite{kamienny_end--end_2022} & Next token prediction & 3Token  &  Sequence to sequence transformer  \\ 
 \hline
 SymbolicGPT\cite{valipour_symbolicgpt_2021}&  Next token prediction &  Point cloud \cite{qi_pointnet_2016} & Sequence to sequence transformer \\
 \hline
 TPSR \cite{shojaee_transformer-based_2023} & Next token prediction & 3Token & Sequence to sequence transformer with MCTS \\ 
 \hline
 DGSR-MCTS \cite{kamienny_deep_2023} & Mutation of syntax tree, Next token prediction  & 3Token &  Sequence to sequence transformer with MCTS \\
 \hline 
 \ours{} (ours) & Select rule from a grammar & Modular & Neural-guided MCTS  \\
\bottomrule 
\end{tabularx}
\end{table}

\subsection{Continual Learning for Equation Discovery}\label{sec:related_work_continual_learning}

The second group of approaches of our interest, and the one closer to our present proposal, attempts to transfer experience across multiple ED tasks. 
One of the first approaches in this direction was \emph{BACON} \cite{langley_scientific_1987}, in which heuristic rules are used to find physical laws. 
Papers published over the past year have attempted to replace these handwritten heuristics with neural models that automatically detect patterns in the data sets. Recently, motivated by its huge successes in natural language processing, transformer-based architectures have been used, framing equation discovery as a sequence-to-sequence problem. The data set is encoded as a text, and the equation is generated token by token. \Cref{table_approaches} gives an overview of how recent approaches define their search space, represent the tabular data, and which algorithm they use.

\emph{NeSymReS} \cite{biggio_neural_2021} uses  large-scale pre-training on generated data. The data sets are encoded with the \emph{Set Transformer} architecture \cite{lee_set_2019}. A beam-search is used to sample candidates from the decoder. The generation of training data is based on the work by Lample and Charton \cite{lample_deep_2019}. Their main focus is on how the structure of the syntax tree can be sampled without biases, e.g. towards too deep or left-leaning trees. Once they sampled the structure, internal nodes and leaves are decorated from a list of possible operators or mathematical entities (integers, variables, constants). \emph{SymbolicGPT} \cite{valipour_symbolicgpt_2021} uses a transformer architecture too, but with a permutation-invariant data set encoder. \emph{PointNet} \cite{qi_pointnet_2016} interprets the data sets as point clouds. 

By combining learning over data sets and MCTS as a search algorithm, \emph{DGSR-MCTS} \cite{kamienny_deep_2023} is the approach most similar to ours. While their approach processes the data set as sequential text, as already described in their previous work \emph{E2E} \cite{kamienny_end--end_2022}, we consider different methods to embed data sets. In their algorithm, the transformer-based architecture makes an initial suggestion for a formula and then suggests $k-$mutations for each node in the search tree. This makes it possible to start with complex equations. Our approach, which starts from a start symbol, offers the advantage that the search space is searched systematically and in the worst case converges to an exhaustive search. 
\emph{TPSR} \cite{shojaee_transformer-based_2023} is also based on \emph{E2E} and tries to combine it with MCTS. In contrast to \emph{DGSR-MCTS}, the main differences are a different reward function and that the neural network is not trained on suggesting mutations, but to complete the partial equation. In addition, rollouts are cached to speed up the search.

\subsection{Domain Knowledge }
All the methods described above are based on next token predictions and therefore use a very simple grammar with the rule $ S \rightarrow T S$. They require a separate check whether the generated formula is syntactically valid at all. The use of a more complex grammar \cite{TodorovskiDzeroski1997,Probabilistic_grammars_for_equation_discovery} offers the advantage that only valid formulas can be created, and domain knowledge can already be incorporated into this grammar. 
\emph{NSRwH} \cite{bendinelli_controllable_2023} extends \emph{NeSymReS} with the ability to include hypotheses as text prompts. These hypotheses can be possible  subtrees like $sin (x_0)$ or symmetries between variables.
For single-task ED, grammars are already used.
\emph{GVAE} \cite{kusner_grammar_2017} maps the rules used from the grammar to create a one-hot representation of the syntax tree. This one-hot representation is then used as input and output of a variational autoencoder.
Brence \emph{et al.} \cite{brence_probabilistic_2021} propose probabilistic grammars for equation discovery but without a guiding neural network. The authors present a theoretical investigation of the expected number of iterations to converge for uniform and ``biased'' grammars. Chaushevska \emph{et al.}
\cite{chaushevska_learning_2022}, in a similar vein to the 
\emph{Machine scientist} \cite{guimera_bayesian_2020},  discuss ways to learn the production rule probabilities from a corpus of known equations.

The work presented in \cref{sec:related_work_continual_learning} uses tabular data sets as direct input to their systems, but an important component, the encoding methods themselves are not analyzed in isolation and are only used and tested in the complete system. To allow comparisons between different methods for embedding the data sets in the following section, we present contrastive learning for tabular data sets. Similar to \emph{dataset2vec} \cite{jomaa_dataset2vec_2021}, we define an auxiliary task that should embed batches from the same data set closer to each other than batches from different data sets. Unlike \emph{dataset2vec}, we sample our batches row-wise rather than cell-wise. Moreover, we assign the same label to batches from equations that differ only in constants.

\subsection{Summary}
In summary, the proposed framework of \ours{} incorporates elements of the above approaches, but does so in being the only system that (i) does continual learning across data sets, (ii) employs MCTS for training and testing, and (iii) uses a grammar to control search. Further, it features a modular architecture that allows to address the research questions from the beginning in a systematic manner. 

\section{Methods}\label{sec:methods}
To give an overview of \ours{} and how its modules interact with each other, we first explain its architecture and training procedure. Subsequently, we will define and discuss the individual modules in detail. 
One of the basic elements of \ours{} is to provide domain knowledge to the system by means of grammars. As in LAGRAMGE \cite{TodorovskiDzeroski1997}, we employ context-free grammars for this purpose (see  \cref{sec:context_free_grammar}). These grammars define the actions which are possible to construct an equation, thereby structuring the search space of the equation discovery task. Each equation is represented as a syntax tree, which is explained in \cref{sec:exploration_of_search_tree}.
Equation discovery is, given a fixed data set, a deterministic Markov decision process. We summarize AmEx-MCTS \cite{amex} in \cref{sec:AmEx-MCTS}, an alternative for Classic MCTS, which visits each terminal node only once but keeps the same guarantees.
Since equation discovery focuses on finding a few outstanding equations rather than many good ones, we use in risk-seeking MCTS  the $\mathit{max}$ operator instead of the $\mathit{mean}$ during the backpropagation step (\cref{sec:modification_mcts}).
In order to compare or pre-train neural architectures for embedding tabular data sets, we introduce contrastive learning for tabular data sets in \cref{sec:contrastive_loss}

\subsection{\ours{}}\label{sec:EquationFinder}

\begin{figure}[pt]
  \includegraphics[width=\linewidth]{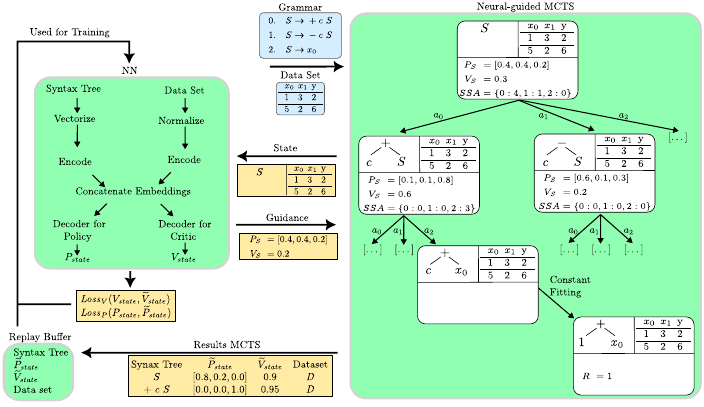}
  \caption{\textbf{Interaction between components in \ours{}.} 
   For each state in the search tree (see the right-hand side of the figure), the neural network gives guidance on which action to choose next ($P$) and how good the current state ($V$) is. Each state has a number of visits ($SSA$), which indicates how often its child nodes are visited. The results of the MCTS are denoted by a tilde and are stored in the replay buffer. When the neural network is trained, a batch is sampled from the replay buffer and the difference between the results of the MCTS and the current prediction of the NN is used to update the weights of the model.} \label{fig:overview_neural_guided_mcts_detail}
\end{figure}

The tree search for the best-fitting equation can be performed by an MCTS. The decision, which child node of the search tree to explore further, depends on a prior. As in the DreamCoder \cite{ellis2021dreamcoder}, a neural network  provides this prior. But in contrast to the DreamCoder, the prior in \ours{} depends on the syntax tree discovered so far and not only on the previous token. \Cref{fig:overview_neural_guided_mcts} shows a high-level overview of the components of \ours{}.
The input into \ours{} is a grammar and a tabular data set. 
Syntax trees are created in an MCTS inspired by AlphaZero \cite{silver2018alphazero}. At each step, the neural network provides guidance which grammar rule should be applied next. If constants are provided in the grammar, these will be fitted at the end. The output of the system are the $k$ best equations that were found. While running the MCTS, results are stored in a replay buffer, which is later used to train the neural network. The structure of the neural network and how to train it is described in the following. 

\subsubsection{Architecture}\label{sec:EquationFinder_architecture}
The neural network for guiding the MCTS has two different information pipelines, as shown in \cref{fig:overview_neural_guided_mcts_detail}. First, the measured values of the experiment as tabular data and second, the current state of the syntax tree. In preliminary experiments, we found that for syntax trees written in prefix notation, \ours{} can predict which rules are possible by using a standard text transformer \cite{vaswani2017attention} to embed the syntax tree. Therefore, no further experiments for syntax tree embeddings are presented here. The encodings of the syntax tree and the data set are concatenated and serve as input to two decoders. The task of the first decoder (policy) is to predict a prior for each rule in the grammar that this rule should be applied next. The other decoder (critic) estimates how good the current state is.

\begin{algorithm}[pt]
\caption{\textsc{searchEquation}}\label{alg:search_equation}
    \begin{algorithmic}[1]
        \Require $\mathrm{model} , \mathrm{data\_set}, \mathrm{grammar}$
        \State $S_0 \gets \textsc{GetInitialState} \left( \mathrm{data\_set} \right) $
        \State $\mathrm{states},\ \mathrm{mcts\texttt{\_}distributions},\ \mathrm{rewards} \gets [S_{0}], [], []$
        \State $i \gets 0$ 
        \While{$\mathrm{states}[i]\mathrm{.equation}\ \mathrm{not}\ \mathrm{done}$}
        \State $\mathrm{mcts\texttt{\_}distributions}[i] = \textsc{RunMCTS} \left( \mathrm{states}[i],\ \mathrm{model} \right) $
        \State $a \gets \textsc{SampleAction} \left( \mathrm{mcts\texttt{\_}distributions}[i] \right) $ 
        \State $\mathrm{states}[i+1], \mathrm{rewards}[i] \gets \textsc{GetNextState}(\mathrm{states}[i], a)$
        \State $i \gets i + 1$
        \EndWhile
        \State \Return $\mathrm{states},\ \mathrm{mcts\texttt{\_}distributions},\ \mathrm{rewards}$
    \end{algorithmic}
\end{algorithm}

\subsubsection{Training}\label{sec:EquationFinder_training}
The idea behind any AlphaZero-related approach is that the guiding neural network and the MCTS reinforce each other. The MCTS visits promising states more frequently by training the policy to predict the results of the MCTS. 
The pseudocode for sampling an equation is given in \cref{alg:search_equation}.

{\ours{} contains two loops: an outer loop that builds the equation for training the guiding neural net and an inner MCTS loop that decides which action to take in each outer loop iteration.  
\Cref{alg:search_equation} describes the outer loop. Starting from an initial state (line 2), the equation is assembled by selecting production rules from the grammar (line 7). To decide which rule to pick from the grammar, the inner MCTS loop is started for each intermediate state in the outer loop (line 5).  \Cref{fig:overview_recognition_module} illustrates the interaction between the MCTS and the guiding NN, when one node of the outer syntax tree is expanded.
For the inner loop, the initial state is the current intermediate state of the outer loop. To estimate the best action to expand the intermediate state, $n$ MCTS simulations are performed. 
 In each simulation step, the NN is queried for guidance. After the simulations, the visit counts $\mathit{Ssa}$ from the intermediate state to its child nodes are evaluated. At test time, the action with the most visits is executed. At training time, based on the visit counts, a distribution is created, and the next rule is sampled from this distribution. Once the outer equation is in a terminal state, the intermediate states, MCTS distributions, and discounted rewards are returned to train the guiding net (line 9).
 The distributions are used to train the policy and the rewards to train the critic subnet of the guiding neural network.
The reward for a syntax tree depends on the following states:

\begin{figure}[tp]
  \includegraphics[width=\linewidth]{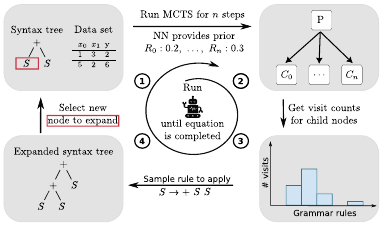}
  \caption{Sequential generation of an equation using the neural network from \ours{}. Based on the visit counts of the MCTS, a probability distribution is calculated to sample which production rule from the grammar should be used to expand the syntax tree. The neural network is trained with the visit counts from the MCTS as well as the reward of the finished equation.}
  \label{fig:overview_recognition_module}
\end{figure}

\begin{enumerate}[label=(\roman*)]
    \item maximum depth exceeded: $r= -1$ 
    \item maximum number of constants exceeded: $r= -1$ 
    \item maximum number of nodes ($node_{max}$) exceeded: $r= -1$
    \item all nodes expanded and $\mathit{MSE}(y_{calc}, \widetilde{y}) > 2: r = -1$
    \item all nodes expanded and $\mathit{MSE}(y_{calc}, \widetilde{y}) < 2: r = 1 -\mathit{MSE} (y_{calc}, \widetilde{y})$
    \item all other cases: $r=0$
\end{enumerate}
Where $\mathit{MSE}$ is the relative root mean square error defined in \cref{equ:relative_mean_square_error}
\begin{equation} \label{equ:relative_mean_square_error}
\mathit{MSE} = \sqrt{\frac{\sum_{i = 0}^{|D|}(y_{i}^{calc} - \widetilde{y}_i)^2}{|D| \cdot}}. 
\end{equation}

$|D|$ denotes the length of the data set, $\widetilde{y}$ the $y$ values from the data set and $y^{calc}$ the $y$ values calculated with the syntax tree of the state.

\subsection{Context-Free Grammar}
\label{sec:context_free_grammar}
We use context-free grammars to define our search space and incorporate existing domain knowledge into the equation discovery process.
A context-free grammar \cite{CHOMSKY1959137} is defined as a tuple $\mathcal{G}=(\mathcal{N},\mathcal{T},\mathcal{R},\mathcal{S})$, where
$\mathcal{N}$ is the set of non-terminal symbols,
$\mathcal{T}$ is the set of terminal symbols,
$\mathcal{R}$ is the set of production rules, and
$\mathcal{S}$ is the start symbol.
Production rules have the form $(A \rightarrow \alpha)$,
where $A$ is a nonterminal ($A \in \mathcal{N}$) and
$\alpha$ is a sequence of terminals and nonterminals ($\alpha \in (\mathcal{N} \cup \mathcal{T})*$).
This production rule states that $A$ can be substituted by $\alpha$.
$\mathcal{T}$ is the set of terminal symbols, i.e., those symbols which cannot be expanded. 
Beginning with a start symbol $\mathcal{S}$, an equation can be constructed by repeatedly applying the rules in $ \mathcal{R}$ until all non-terminals are resolved to terminals. This equation can be represented as a syntax tree as explained in \cref{sec:exploration_of_search_tree}. 
The grammars used in our experiments are given in the appendix. While we are using basic operations such as plus and power in our grammar, complex expressions such as polynomials or even neural networks could be added as rules to the grammar. 

A grammar not only describes the structure of expressions in a language, it can also be used to sample expressions in that language. If each production rule is given a probability of being selected, a probabilistic context-free grammar (PCFG) is created (see, e.g., the use of PCFGs in the ProGED system \cite{brence_probabilistic_2021}). These probabilities can be assigned by, e.g., an expert or a neural network.

\subsection{Exploration of the Search Space}\label{sec:exploration_of_search_tree}
We represent our equations as syntax trees. 
A syntax tree \( \mathcal{S} \) consists of nodes \( \mathcal{N} \) and edges \( \mathcal{E} \). 
The edges connect two nodes $n_i$ and $n_j $ where $ n_i , n_j \in \mathcal{N} \wedge i \not = j$.
To represent an equation, we use a syntax tree in which the inner nodes are operators and functions, and the leaf nodes are variables and constants. 

Since our goal is to find equations, we need a systematic way to explore the space of possible equations. For this, we use a search tree in which the edges are the chosen grammar rules, and the nodes are states consisting of a syntax tree and the current data set. The task of finding the best equation for a given data set can be formulated as the following sequential decision-making problem: Beginning with the start symbol $\mathcal{S}$, which rules from the grammar do we have to apply to obtain the equation that best fits to our problem?

The leaves of the search tree can be in one of three conditions:

\begin{enumerate}
    \item States whose syntax tree is not completed yet, and the child states are not explored.
    \item States that will not be further explored because their syntax tree violates some constraints (e.g., the maximal number of nodes in the syntax tree).
    \item States which are terminal, and their syntax tree represents a complete equation.
\end{enumerate}

The search for the best fitting equation can be done by a Monte-Carlo Tree Search (MCTS) \cite{kocsis_bandit_2006,coulom_efficient_2007}.
MCTS consists of four basic steps: \textit{selection}, \textit{expansion}, \textit{simulation} and \textit{backpropagation}.

The first step of every MCTS cycle is the selection. Starting at the root, a decision has to be made which path to explore further, i.e., which grammar rule to apply to the syntax tree within the search tree node. This is repeated at every node, until a leaf node is found.
The expansion step is performed, when a node with a syntax tree in condition 1 is reached. It essentially boils down to applying a grammar rule to the current syntax tree, and by doing this adding a new node to the search tree.
The simulation step generates a value that estimates how good the partial equation suits the data set. How this value is determined depends on the apporach: a random roll-out can be carried out or an external oracle can evaluate the current state.
During backpropagation, this value is propagated up the search tree back to the root.

Since we aim to build a system which, among other things, fulfills Requirement 1 (Using previous searches), we do not use plain MCTS, but build upon the AlphaZero framework \cite{silver2018alphazero}. The key idea of AlphaZero is to use MCTS to learn a good decision-making rule -- also known as policy -- and to use this policy to guide the MCTS into promising regions of the search tree. MCTS and policy should therefore support and improve each other.

For the four MCTS steps described before, using the AlphaZero framework has the following consequences: In the selection phase, we use the ``predictor upper confidence bounds applied to trees'' (PUCT) formula \cref{equ:PUCT} to select the grammar rule to apply. This formula balances, with a hyperparameter $c$, exploitation (applying the rule that looks most promising) and exploration (trying other rules that might be better) by taking a couple of factors into account: How often the current node $s$ was visited during the search $|S|$ (visit count of a node), how often the grammar rules were applied from this node $|\textit{Ssa}|$ (visit counts of the edges), the values of the child nodes gained during the simulation steps $Q(s,a)$ (Q-values), and an initial prior $P(s,a)$ for the action $a$ (potentially a uniform distribution over the grammar rules). In the end, the action which maximizes the PUCT score is chosen. 
\begin{equation}\label{equ:PUCT}
    \mathit{PUCT}(a) = Q(s,a) + c \cdot P(s,a) \cdot \frac{\sqrt{|S| + 1}}{|\textit{Ssa}| + 1}
\end{equation}
Further, in the simulation step, the value of a state (syntax tree and data set) in the search tree is calculated by a \textit{critic}, which is a sub-network of the neural network guiding the search. The training of the critic and a policy was  already described in \cref{sec:EquationFinder_training}.
In \cref{sec:AmEx-MCTS}, we describe how we modified the MCTS algorithm, so that each leaf node is visited only once, enhancing the efficiency of MCTS in equation discovery.

\subsection{AmEx-MCTS}\label{sec:AmEx-MCTS}
For our approach \ours{}, MCTS is the search backbone.
Classic MCTS, however, comes with a couple of drawbacks. 
Depending on the threshold between prediction and real data at which an equation receives a reward $> r_{min}$, we can set the sparsity of rewards in the search tree. If the threshold is low, the MCTS receives only very few signals as to which regions of the search tree should be examined more closely. If the threshold allows larger variations, there is a risk that the MCTS will get trapped in local optima or just waste a lot of computation without gaining new information by visiting these local optima over and over again.
The equation discovery task features a large deterministic search spaces with a high branching factor, but since we are interested in short, concise equations, the individual branches are not very deep.
Especially in such scenarios, Classic MCTS can get trapped in previously mentioned local optima.
To counteract, we make use of our MCTS extension AmEx-MCTS~\cite{amex} within the learning setup.

Within the MCTS algorithm, the visit counts on the nodes (states) and edges (state--action pairs) are the most important values. The visit counts of the edges starting at the root node of the search are usually used to calculate the action probabilities for the next step.
The idea behind AmEx-MCTS is to avoid revisiting fully explored subtrees by keeping track of them, but to still keep the visit counts as they would be when using Classic MCTS. 
To do so, we decouple the value updates, visit count updates, and the selected path in the backpropagation step.

As visualized in \cref{fig:amex}, we do so by adapting the two steps \textit{selection} and \textit{backpropagation}. Instead of directly selecting the action with the highest score $a_{max}$, we select the action $a_{select}$ with the highest score that does not lead to a fully explored subtree. Nevertheless, the visit count of $a_{max}$ is increased. 
In order not to bias the value estimates in higher nodes of the tree, we only propagate the reward achieved by following $a_{select}$ further up the tree if it is greater than the current value estimate of the child node reached by selecting $a_{max}$.
For pseudocode and further details, we refer to our original manuscript \cite{amex}.

\begin{figure}[tp]
    \centering
    \def\svgwidth{\linewidth}
    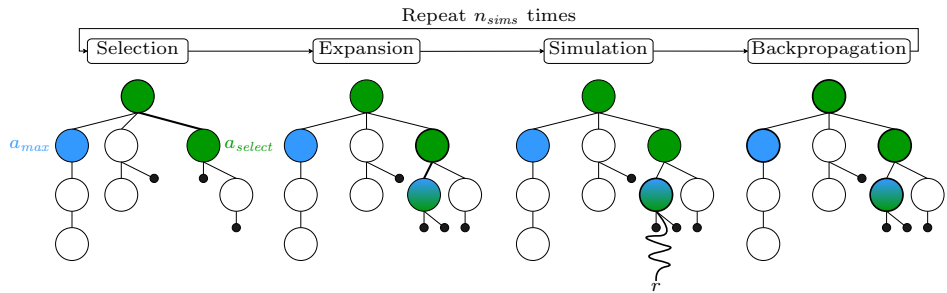
    \caption{\textbf{Improving MCTS by ignoring already explored subtrees and leaves by focusing on the unknown.} Updating the search strategy within MCTS by separating ``incrementing visit counts'' (displayed in blue) from the selected child nodes (displayed in green) to explore more while keeping the number of iterations \(n_\mathit{sims}\) the same.}
    \label{fig:amex}
\end{figure}

\subsection{Risk-seeking MCTS} \label{sec:modification_mcts} 
As Petersen \emph{et al.}\cite{Deep_symbolic_regression_Recovering_mathematical_expressions_from_data_via_risk_seeking_policy_gradients} already pointed out, a search in the domain of equation discovery is only supposed to find a small number of well-matching equations. If a branch of the search tree leads to only one good result, and the other paths in that branch lead to incorrect equations, the path to the single good equation should not be affected by the others. Therefore, we use the max operator instead of the mean operator when updating the Q-values in the backpropagation step of the MCTS. 
Consequently, the Qsa-values near the root node increase faster than would be the case with using the mean during backpropagation. To mitigate the effect of the Qsa-values and keep exploring, we use for the PUCT-Formula \cref{equ:PUCT} $c=10$.  

Since the first rules used are of particular importance and can only be corrected with difficulty later in the search tree, more simulations in the MCTS are used for the first decisions. 
The number of simulations is based on the number of nodes ($|\text{Syntax Tree}|$) in the syntax tree and is calculated by
\begin{equation}
sim_\mathit{MCTS}= \max \left( sim_{init} \cdot 4^{- (|\text{Syntax Tree}| -2) }, 10 \right).
\end{equation}
The -2 in the exponent is due to the fact that each syntax tree initially has a $y$ and a start node. 
In AlphaZero \cite{silver2018alphazero}, the simulation restarts at the root node after each expansion step. 
In our experiments, however, a complete path of the search tree is explored in each simulation. By doing this, we receive a reward at the end of each path and are not solely dependent on the critic subnet of the guiding NN to determine the quality of an intermediate node.  
In equation discovery, overlong search paths can be prohibited as the size of the syntax tree increases with each action taken, and we can set a maximum size of the tree.

\subsection{Contrastive Learning for Tabular data sets}\label{sec:contrastive_loss}

\begin{figure}[tp]
  \includegraphics[width=\linewidth]{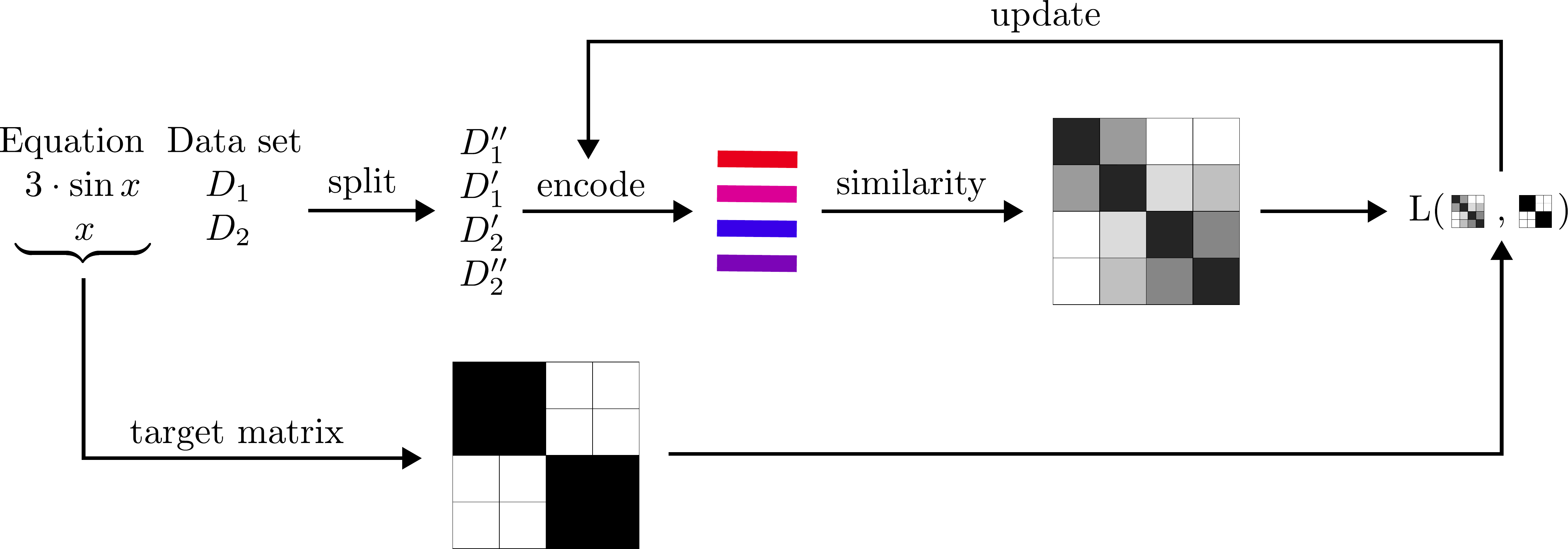}
  \caption{Visualization of the contrastive loss between data sets for different equations.}
  \label{fig:overview_contrastive_loss}
\end{figure}
In our preparatory experiments, we found that embedding the tabular data sets is one of the main challenges for \ours{}. We introduce contrastive learning for tabular data sets to compare the ability of different embedding architectures in isolation.

The idea behind contrastive learning in general is that the model should learn to map two examples from the same class (positive samples, e.g., two images of dogs) closer to each other in the latent space than two examples from different classes (negative samples, e.g., images of a dog and a cat). 
The equation for contrastive loss \cite{khosla_supervised_2020} is given by:
\begin{equation}
    L_{self} = - \sum_{i \in \displaystyle{I}}^{} log \frac{exp [\left\langle \textbf{z}_i, \textbf{z}_{j(i)} \right\rangle]}{\sum_{a \in A(i)}^{} exp [\left\langle \textbf{z}_i, \textbf{z}_{a} \right\rangle] } , 
\end{equation}
where $\textbf{z}_i$ is the sample working as anchor. 
$\textbf{z}_{j(i)}$ denotes the positive example to the anchor. $\textbf{z}_a$ are all samples in the batch, excluding $\textbf{z}_i$. \\

The concept of contrastive learning is known from the image domain, where the positive example of an image is created by manipulation such as zooming, cropping, or rotation. Tabular data, on the other hand, already consists of $n$ individual rows, which were all generated by the same process. 
We therefore sort the rows according to the y-values and split each table into two halves. We sort the values first to force the net to learn the mapping of parts of functions together and not just calculate statistics like the mean.  Subsequently, the tables are processed by the data set encoder. The cosine similarity is calculated between all embeddings $s$ in the batch. By a transformation with $\nicefrac{x+1}{2}$, the cosine similarity is mapped into the value range $[0, 1]$. A value of 0 indicates that the embeddings point in opposite directions, while a 1 means that the embeddings point in the same direction. 
The similarity value in the target matrix $\mathbf{t}$ should be 1 for the split data sets that originate from the same data set, i.e., $t_{i,i}, t_{i,i+1}, t_{i+1,i} t_{i+1,i+1} = 1$. To reference these elements, we will use the subscript $\mathit{self}$, for the other elements in the batch, we use the subscript $\mathit{other}$, and the values in the target matrix are set to 0. 
At training time, when the original equations for the data sets are known, all split data sets in the target matrix with the same original equation, apart from constants, are set to 1. We proceed in this way because the structure of the syntax tree is independent of the value of its constants.

The contrastive loss for data sets $L_\mathit{contrastive}$ is defined as 
\begin{equation}
    L_\mathit{contrastive} = \mathit{BCE}(\mathbf{s}_\mathit{self}, \mathbf{s}_\mathit{self}) + \lambda \cdot \mathit{BCE}(\mathbf{s}_\mathit{other}, \mathbf{s}_\mathit{other}),
    \label{equ:contrastive_loss}
\end{equation} where $\mathit{BCE}$ is the binary cross entropy.

With $\lambda$, we can trade off whether it is more important to place split data sets from the same source close together in the embedding space or to separate data sets originating from different data sets further. 
Beyond the scope of our analysis is how an individual number from a data set should be represented (direct use, as bit encoding, as text, etc.) \cite{gorishniy_embeddings_2022}.

\section{Generation of Training Data Sets}\label{sec:generation_of_training_data_sets}
As a system that learns continuously, \ours{} requires training sets for pre-training. In this section, we explain the way data sets are generated for the pretraining of the system.

As Ellis \emph{et al.} \cite{ellis2021dreamcoder} have shown, grammars and the inclusion of concepts shared by many solutions are a suitable way to generate complex training data.
While the ability to add concepts will be added in future work, we now only use fixed grammars.
We use three grammars in total, called Grammar A, Grammar B, and Grammar C in the following (see the Appendix).

Grammar A  produces various equations used to train and test \ours{} and the ability of its tabular data set embedding architectures.
This grammar includes, among others, polynomials with a degree up to 4, trigonometric, inverse, logarithmic, and exponential functions with two variables.
Grammar B and Grammar C are designed to define a search space for equations in the style of the Nguyen equations and are used to compare DSO \cite{Deep_symbolic_regression_Recovering_mathematical_expressions_from_data_via_risk_seeking_policy_gradients}, AmEx-MCTS, and Classic MCTS. The key differences between them are as follows: In Grammar B, $x_0^i, i \in \left[2,3,4,5,6 \right]$ are included as rules, so they do not need to be built by multiplying $x_0$ several times.
Additionally, the inner functions in  $sin$, $cos$, and $log$ are not recursive, allowing only functions with one operator.
In total, Grammar B should be expected to be more suitable for the Nguyen equations than Grammar C. 

The following describes how a grammar can be used to generate training data.
Based on a grammar, syntax trees can be sampled, and for each valid syntax tree, a table of the form $x_0$, $x_1$, and $y = f_{syntax tree}(x_0, x_1)$ is generated. We only allow syntax trees smaller than 25 nodes and up to 5 constants in the value range $\left[ 0.5, 5 \right]$. In addition, the depth of the syntax tree is limited to 10.

The interval boundaries $x_{i}^{min}$ and $x_{i}^{max}$ for sampling are set to -5 and 5, respectively.
If an $x_i$ is used in a logarithm, $x_{i}^{min}$ is automatically set to 0. If $x_{i}^{max}$ - $x_{i}^{min} < 2$, both values are resampled.
If all constraints are satisfied, 100 values between $x_{i}^{min}$ and $x_{i}^{max}$ are sampled for each $x_i$. These sampled values are evaluated on the generated syntax tree, and if all rows in the table are valid, the data set is used. If an error in the evaluation occurs, the process is repeated from the beginning.

\section{Experiments} \label{sec:experiments}
In this section, we discuss three different experimental setups\footnote{Code publicly available: \url{https://github.com/wwjbrugger/EquationFinder}} to answer our research questions.
The most important parameters are given in the appendix in \cref{tab:hyperparameter}, and all parameters are given in the repository.
We switch the order of answering the questions and start with \textbf{R3}, because the choice of the data set encoding can then be fixed and used in all other experiments. 
In Experiment~1, we introduce different NN architectures for encoding tabular data sets and compare them using contrastive loss (see Section \ref{sec:contrastive_loss}). 
In Experiment~2,  these NN architectures are components of \ours{} and are trained and tested on a set of handcrafted equations to show the effects of different learning methods. 
The effects of supervised learning, the number of MCTS simulations during training, and the use of only one encoding (syntax tree or tabular data set) are then investigated (\textbf{R1}-\textbf{R3}). 

In the experiments, we use AmEx-MCTS as our main search method. In Experiment~3, we validate this approach by comparing AmEx-MCTS with Classic MCTS for the Nguyen data sets and demonstrating the benefits of using AmEx-MCTS (\textbf{R2}). Further,  we show in this experiment the advantage of using grammars to incorporate domain knowledge in neural-guided equation discovery instead of using a token-based approach, such as DSO  \cite{Deep_symbolic_regression_Recovering_mathematical_expressions_from_data_via_risk_seeking_policy_gradients} (\textbf{R4}).
We finally discuss the research questions \textbf{R1} to \textbf{R4} in light of the three experiments in Section 6.4.

\subsection{Data Set Embeddings}\label{sec:experiments_data_set_embeddings}
The contrastive loss for tabular data sets introduced in \cref{sec:contrastive_loss} is used to compare different architectures for embedding tabular data generated from Grammar A. We set $\lambda $ in  \cref{equ:contrastive_loss}, which balances the importance of embedding similar data sets close to each other versus separating dissimilar data sets, by default to 0.1.

An MLP is used as a baseline of a non-permutation-invariant architecture with regard to rows or columns. 
Using LSTMs \cite{LSTM_Hochreiter}, an architecture is tested that processes the tables row-wise. 
A variation of LSTM is examined with Bidirectional-LSTM (Bi-LSTM), in which the data set is input forward and backward into an LSTM, and the result of both directions is concatenated. 
Non-Parametric Transformers (NPTs) \cite{kossen_self-attention_2021} is a permutation-invariant transformer architecture which is based on the Set Transformer \cite{lee_set_2019}. 
Kamienny \emph{et al.} \cite{kamienny_end--end_2022, kamienny_deep_2023} use a transformer architecture inspired by transformers for sequential texts. In the subsequent discussion and figures, we refer to this aproach as Text Transformer.

\begin{figure}[tp]
  \centering
  \includegraphics[width=0.7\linewidth]{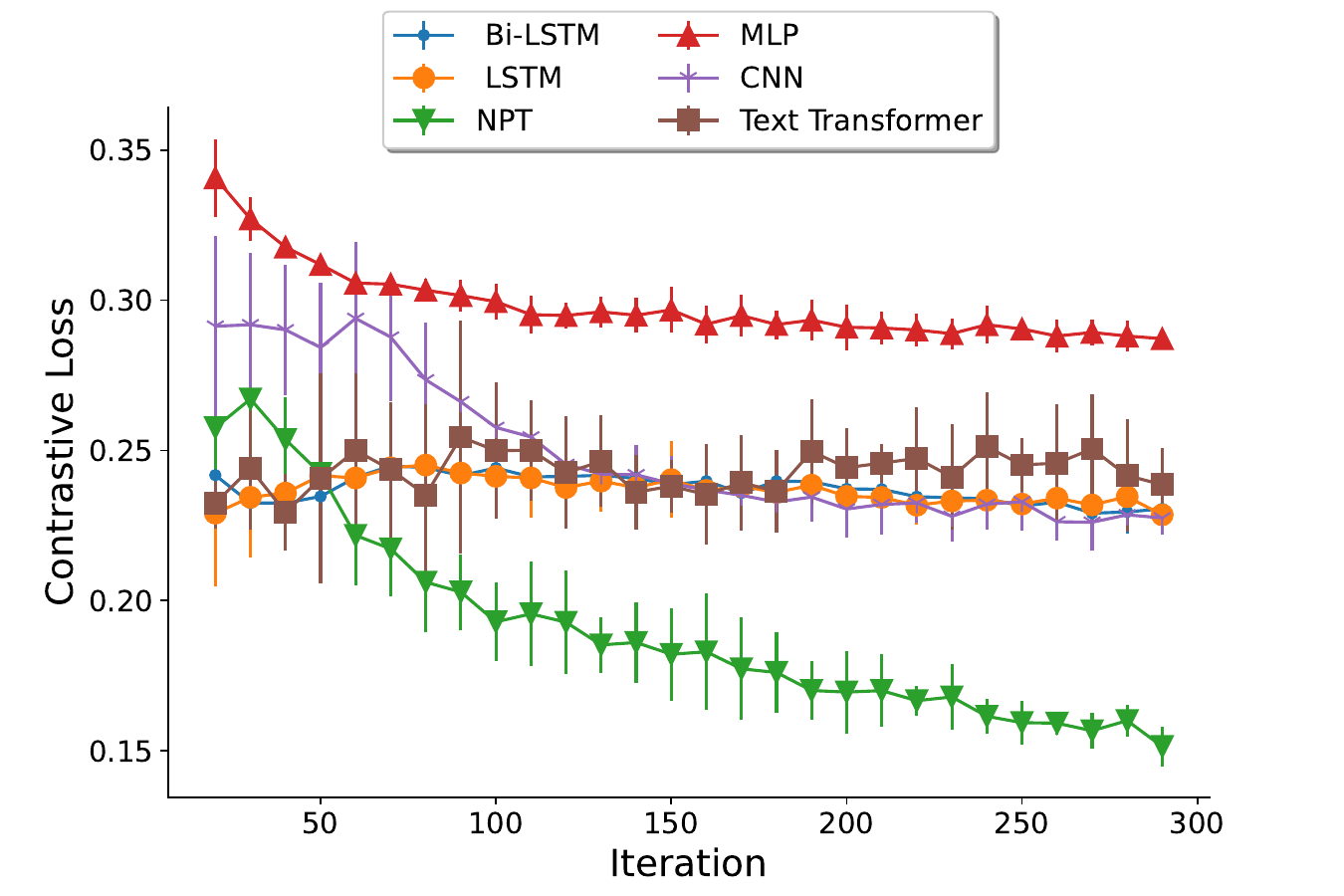}
  \caption{\textbf{Contrastive loss for tabular data set embeddings:} NPT, LSTM, CNN, Text Transformer and MLP are compared. Mean value and standard deviation are calculated for 3 seeds.}
  \label{fig:contrastive_loss}
\end{figure}

Inspired by the DreamCoder \cite{ellis2021dreamcoder}, we plot $x_{i}$ against $y$ and concatenate the resulting plots as channels. A convolution-based architecture is used as the embedding architecture. While plotting the data makes them permutation-invariant, this representation cannot preserve the relationship between $x_0$ and $x_1$.
In the subsequent discussion and figures, we refer to this approach as CNN.

Since neural networks do not tolerate extreme values as input, the $y$ values from the data set are scaled with $y_{\mathit{input}} = \nicefrac{y}{max(|y|,1)}$, before they are processed by the NN. 
The scaled values are only used within the NN. For the calculation of the rewards, the original values from the data set are always used. 

Each approach is trained for 300 iterations and the results are shown in \cref{fig:contrastive_loss}. 
The permutation-tolerant architecture of NPT shows the best results with the lowest contrastive loss. LSTM, Bi-LSTM, CNN, and the Text Transformer all achieve a similar contrastive loss. The embeddings of the tables by an MLP have the highest loss. 
The second result is that the Text Transformer is the only method that exhibits overfitting. An explanation could be that this architecture has the most parameters, as \cref{num_trainable_parameter_model} shows, or that the representation of numbers as triplets (sign, mantissa and exponent) leads to this effect.
\begin{table}[tp]
\caption{\label{num_trainable_parameter_model} \textbf{Number of trainable parameters in the used models}}
\begin{tabular}{>{\centering\arraybackslash}p{0.4\textwidth}>{\centering\arraybackslash}p{0.3\textwidth}>{\centering\arraybackslash}p{0.3\textwidth}} 
\toprule
Data set encoder    & Data set encoder only    &  Complete model    \\
\midrule
MLP                 &          71\ 552                               &        144\ 717                               \\
LSTM             &           17\ 408                              &          82\ 381                             \\
Bi-LSTM             &        34\ 816                                 &     44\ 400                                  \\
CNN               &             52\ 312                           &              113\ 189                         \\
NPT                 &          11\ 979\ 054                              &        12\ 037\ 627                                \\
Text Transformer    &             41\ 424\ 384                            &      41\ 546\ 701                                 \\

\bottomrule
\end{tabular}

\end{table}

\subsection{Neural-Guided Equation Discovery } \label{sec:experiments:neural_guided_equation_discovery}

We use Grammar A (see appendix) to measure the influence of the training method, the data set embedding, and the syntax tree embedding. The grammar is selected so that the generated equations cover a broad spectrum of forms. In total, 384 syntax trees with two variables can be generated. A tabular data set with $n=100$ entries is sampled for each syntax tree. The variables $x_0$, $x_1$, and constants are randomly sampled so that an equation can occur multiple times, but each problem is unique.

For data set embedding, the architectures MLP, LSTM, Bi-LSTM, CNN, NPT, the Text Transformer, and no embedding are tested. There are two options for the tree embeddings: either the syntax tree in prefix order is embedded by a transformer architecture, or the embedding is omitted. The training can be performed supervised or with the distribution from the results of the MCTS. The number of simulations allowed in MCTS influences the prediction of the NN, which is intended to guide the search. MCTS with 10, 50, 125, or 250 simulations are examined to show this. The labels used in the supervised training are the actions to build the ground truth syntax tree used to create the tabular data set.

The training covers 200 iterations with 50 problems per iteration. In the first 10 iterations, the NN is not updated. After this cold start,  at the end of each iteration, the neural network is updated 20 times with batches of size 64 sampled from the experience replay buffer. For paths whose final equation leads to a reward smaller than $-0.9$, a uniform distribution is added to the replay buffer instead of the distribution based on the visit counts of the child nodes. The test data set consists of a further 500 randomly generated problems.
\begin{table}[tp]
  \caption{\textbf{Mean number of simulations until equation with $\mathbf{r > 0.999}$ is found for 500 random equations sampled with Grammar A}. In table (a), the syntax tree embedding is an input to \ours{}. In table (b), it is not an input. All learning setups benefit from the tabular data set as input. The approaches that were trained in a supervised manner find a good equation faster. The MCTS-based learning setups does benefit from more simulations.
We underline the best architecture for each learning setup and write the best learning setup for each architecture in bold.\\}
  \label{tab:500_generated_equations}
  \includegraphics[width=0.95\linewidth]{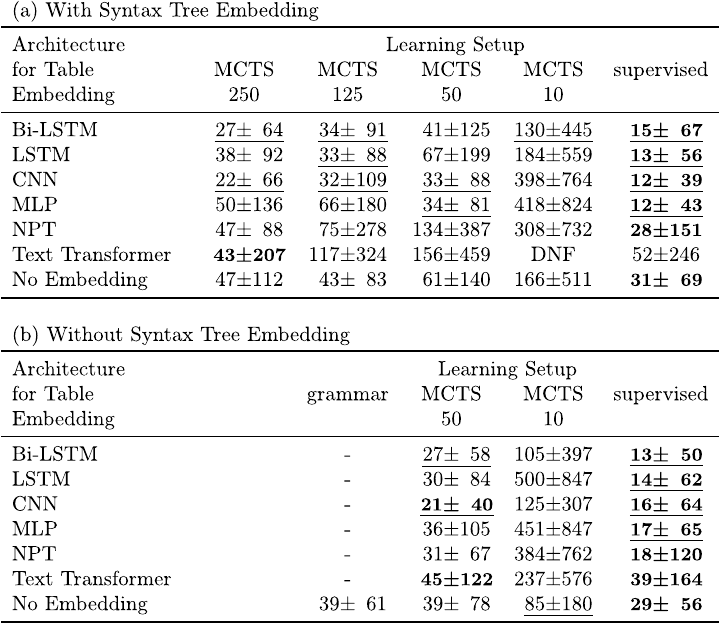}
\end{table} 

\Cref{tab:500_generated_equations} shows for the test set the average number of simulations required to find an equation with a reward $>0.999$.  
The supervised trained models require on average fewer simulations than the models trained with MCTS. The MCTS-based models benefit from more simulations during training. It is noteworthy that, contrary to our experiments with contrastive loss, it is not the NPT that shows the strongest performance but the MLP, CNN and LSTM approaches. 

The models trained without syntax tree embedding show similarly good results as those with the current syntax tree as input. One reason for this is that currently impossible actions are masked when building the tree. In a later analysis, we consider the prior of the guiding network and see that the models with the syntax tree as input actually use this information.
\Cref{fig:histogram_sim_to_perfect_fit} shows histograms of the number of simulations until an equation with a reward of $> 0.999$ is found for the test set. Only a few problems need more than 300 simulations for approaches trained with 250 MCTS simulations (as marked by a grey vertical bar in Figure \ref{fig:histogram_sim_to_perfect_fit}) (2)  or supervised learning (3). The methods without syntax tree embedding (4), uniform distribution (4), and tabular data set (5) show more outliers. The model trained with only 10 MCTS simulations per state has the most outliers (26).
This behavior can be particularly undesirable in the area of equation discovery, where a longer average running time can be acceptable and preferred to a shorter average running time and the risk of not finding a solution at all. 

\begin{figure}[tp]
\centering
  \includegraphics[width=\linewidth]{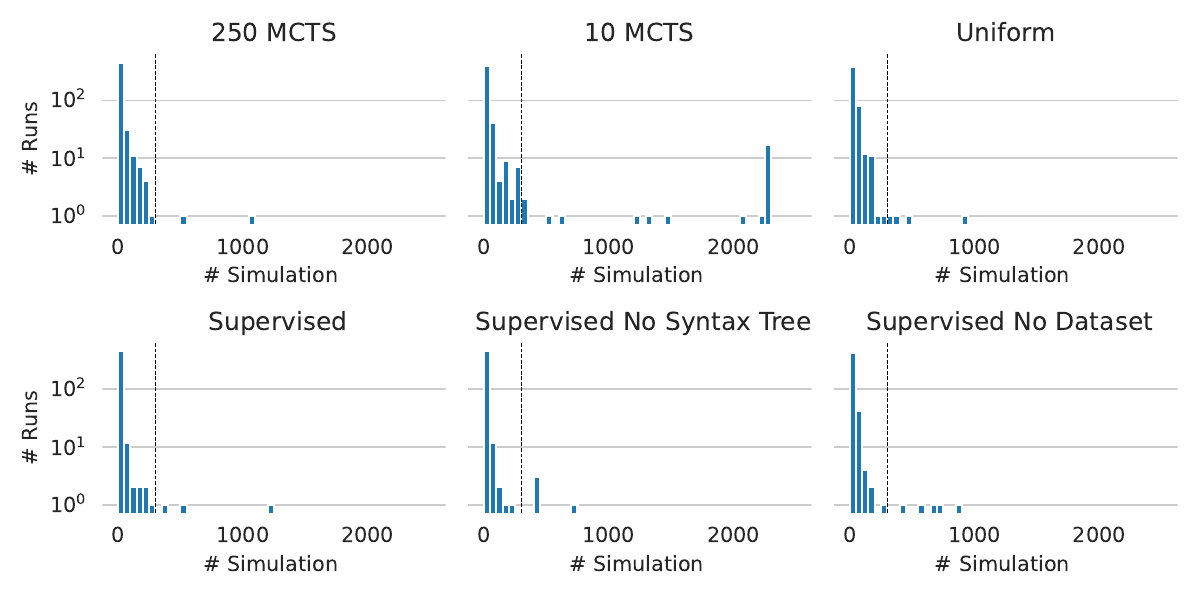}
  \caption{\textbf{Histograms  of number of simulations until equation with $r>0.999$ is found for 500 random equations.}  
  If not otherwise stated, the models embed the tabular data set with the Bi-LSTM architecture.   All approaches find the majority of equations within a few simulations.   The black dotted line marks 300 simulations. While approaches trained with supervised learning or 250 MCTS simulations per state have only a few outliers (for our illustration defined to have more than 300 simulations), the other methods show more outliers.
  In a bigger search space, high values correspond to an equation that could not be found.}
  \label{fig:histogram_sim_to_perfect_fit}
\end{figure} 

To better understand the effects of different learning setups, the priors predicted by the neural guidance for the start node (\cref{fig:priors_start_node}) and an intermediate node (\cref{fig:priors_intermediate_node}) are shown for data sets corresponding to different equations. The actions from 1 to 27 on the x-axis correspond to the rules from Grammar A. Each point in a subplot represents the prior assigned to an action given a data set. If multiple data sets in the test data set are sampled from the same equation, the average and the standard deviation of the priors are provided.

\begin{figure}[tp]
  \includegraphics[width=\linewidth]{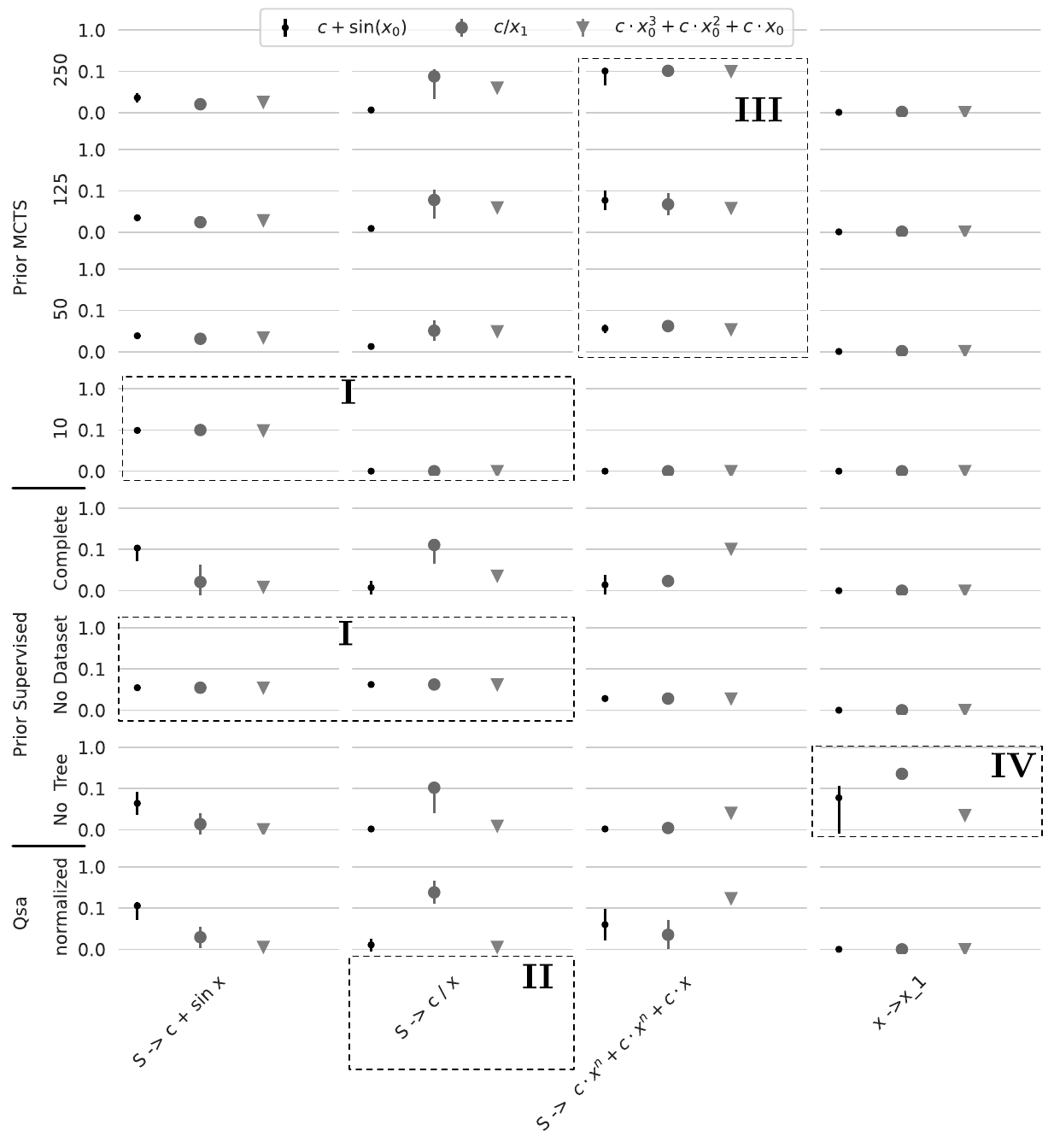}
  \caption{\textbf{Average priors for the initial state for 3 equations} ($c + sin(x_0)$, $\nicefrac{c}{x_1}$, , $cx^3 + cx^2 + cx$). Each row shows the priors of one model for a selection of rules from Grammar A.
    Box $\mathbf{I}$ shows that without data set embedding or when the guiding net is trained with the results of 10 MCTS simulations, the priors cannot adapt to the problem. Column $S \rightarrow c / x$ ($\mathbf{II}$) indicates that the other setups can adapt. 
Box $\mathbf{III}$ illustrates that the guiding nets trained with MCTS results only vary slightly for 50 simulation steps or more. When the current syntax tree is unavailable, impossible rules in the given state get a high prior, as in box $\mathbf{IV}$.       
  When a tabular data set embedding is used, it is calculated by the Bi-LSTM architecture.
  For better readability, we replace $\textit{Variable}$ by $x$ in the rules of Grammar A and use the infix notion.}
  \label{fig:priors_start_node}
\end{figure} 

In \cref{fig:priors_start_node}, the priors for the initial state of the search tree are shown for three different equations. The bottom line shows the normalized Qsa values for a fully explored search tree, representing the ground truth of how good an action is. The approaches trained with the results of the MCTS search aim to approximate this distribution. The priors of the neural network trained with the MCTS performing 250, 125, and 50 steps during training differ only marginally from each other. The predictions between equations vary but less strongly than the normalized Qsa values. One reason could be that we train the policy of the neural network to predict the normalized visit counts of the child nodes. Using $c=10$ in the PUCT equation (\cref{equ:PUCT}) gives the initial prior a high influence on which child node to visit next. As long as the results following an action with a high prior are not bad, the action will receive  many visits, and the network is not forced to change its prediction for this state.

From our experiments, 50 simulations in the MCTS are sufficient to avoid a feedback loop where the MCTS and neural network adversely reinforce each other. With only 10 MCTS simulations, the training fails because the MCTS needs more simulations to explore actions thoroughly, and it always first explores those actions that the neural network suggests.
Our architecture can learn very different priors for different equations, as illustrated by the row for supervised training, which uses the data set and the syntax tree as encoding. When no data set is given, \ours{} cannot distinguish the problems to solve, and the prior for supervised training without data set embedding is always the same.

Figure \ref{fig:priors_intermediate_node} shows the priors for an intermediate node $c + \sin(\textit{Variable})$ for the equations $c + \sin(x_0)$ and $c +(\sin\ x_1)$. Action 26 substitutes $\textit{Variable}$ with $x_0$ and Action 27 with $x_1$.
With this, we test whether the guiding network can merge the information from the data set and the syntax tree. With the exception of the approaches without the embeddings of the tabular data set or the syntax tree, the other approaches are able to do this. In the row for the supervised approach without a syntax tree, it is easy to see how the model also tries to give action 2 $ S \rightarrow c + \sin (\textit{Variable})$ a high probability, as it cannot distinguish in which state the syntax tree is in. The model trained with an MCTS performing  250 simulations per state gives impossible actions a prior greater than 0. 
This is not the case for the approaches trained with 125 and 50 simulations (results not shown).

\begin{figure}[tp]
  \includegraphics[width=\linewidth]{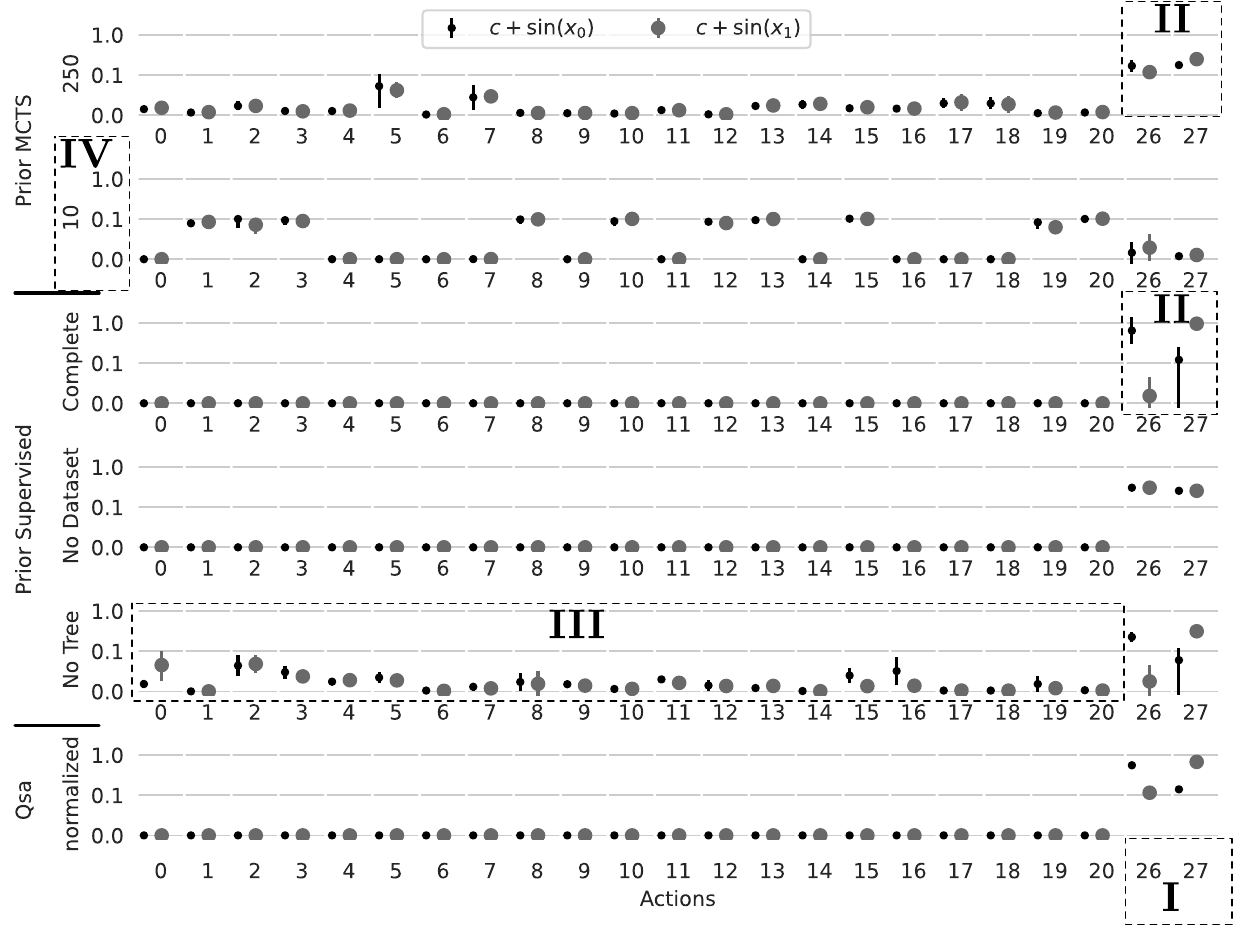}
  \caption{\textbf{Average priors for an intermediate state}. The intermediate state is $c + \sin \textit{Variable}$ for the equations $c + \sin(x_0)$ and $c + \sin(x_1)$. The only actions possible in this state are 26 and 27 (box $\mathbf{I}$). Boxes ($\mathbf{II}$) show that supervised learning and the approach trained with the MCTS results learn different distributions. The reason is that supervised learning is trained with one hot labels, while the MCTS results used for training (see Qsa row) show that both variables are feasible fits. Box $\mathbf{III}$ demonstrates that when the current syntax tree is unavailable, impossible rules in the given state receive a prior $\neq 0 $. The net trained with the results of an MCTS with 10 simulations learns no meaningful prior (Box $\mathbf{IV}$). When a tabular data set embedding is used, it is calculated by the Bi-LSTM architecture.}
  \label{fig:priors_intermediate_node}
\end{figure}

\subsection{Grammars and AmEx-MCTS vs. Classic MCTS}
Equation Discovery is usually considered a deterministic search with the goal of finding a few well-matching equations. To investigate MCTS variants for this task, we propose AmEx-MCTS and risk-seeking MCTS as an adaptation of the classical MCTS.
To show the efficiency of AmEx compared to Classic MCTS, we run both versions for Grammar B and C  on the 12 Nguyen equations \cite{uy_semantically-based_2011} with 5 seeds each.  We used the Nguyen data sets  as reported by Petersen \emph{et al.}\cite{Deep_symbolic_regression_Recovering_mathematical_expressions_from_data_via_risk_seeking_policy_gradients}.

As \cref{tab:comparision_grammar_MCTS_} shows,  AmEx-MCTS requires fewer MCTS-simulations on average until an equation with $r>0.999$ is found than Classic MCTS. At the same time, the number of explored states is higher. This is because Classic MCTS revisits local optima instead of exploring new regions. 
As an example of a token-based approach, we have also added the results of DSO to  \cref{tab:comparision_grammar_MCTS_}.
To compare \ours{} with DSO, we set the number of equations tested by DSO equal to the number of simulations required by \ours. The average running time of DSO is shorter than \ours. In our system, a bottleneck is that all simulations are executed sequentially, whereas they could be partially parallelized. DSO requires a factor of 10 more simulations than the grammar-based approaches. This is because with the grammars, only syntactically correct equations are generated, additionally, with the grammars, domain knowledge is added to the search. 
The experiments in Derstroff \emph{et al.} \cite{amex} show that the risk-seeking approach has slight advantages over Classic MCTS.

{
\setlength{\extrarowheight}{1em}
\begin{table}[tp] 
\caption{\label{tab:comparision_grammar_MCTS_} \textbf{Comparison of MCTS: Classic vs. AmEx and Grammar: B vs. C}  The average is calculated with 5 seeds for the equations from the Nguyen dataset (see the appendix for the equations and the individual results)}
\centering

\begin{tabular}{ccccc}
\toprule
& \makecell{ $\varnothing ~ \text{MCTS-simulation}$ \\ until perfect fit} & \makecell{ $\varnothing \  \text{Explored states} $\\ until perfect fit} & \makecell{ Runtime \\ $[ \text{sec} ]$  }& \makecell{Unsuccessful \\ fits }\\
\midrule
\makecell{Classic MCTS + \\ Grammar B}   &             18149 &             40882 &    298 &                 3 \\
 \makecell{AmEx-MCTS +\\ Grammar B}      &            13071 &             46930 &    326 &                 3 \\
 \makecell{Classic MCTS + \\ Grammar C}  &              12981 &             73184 &    426 &                 6 \\
\makecell{AmEx-MCTS + \\ Grammar C }     &             11483 &             81790 &    422 &                 4 \\
DSO                                      &            177413 &                 - &    105 &                14 \\
\bottomrule
\end{tabular}
\end{table}
}

\subsection{Discussion of the Initial Research Questions}
Based on our research questions, we will discuss and summarize our experiments using the \ours{} framework. The first question, \textbf{R1}, examines whether a search guided by neural networks provides better results than one without.
Our results in \cref{tab:500_generated_equations} corroborate findings by other groups \cite{biggio_neural_2021, kamienny_end--end_2022, valipour_symbolicgpt_2021, shojaee_transformer-based_2023, kamienny_deep_2023} that it does. 

The second question, \textbf{R2}, asks for the best way to train the guiding neural network and how MCTS can be adapted to the domain of equation discovery. 
Although Classic MCTS can be improved with AmEx-MCTS and risk-seeking MCTS,  our results in \cref{tab:500_generated_equations}  support the use of supervised training.
Additionally, the longer running time of MCTS-based approaches must be considered. Since the results of the MCTS are used to train the neural network, and the neural network determines the prior of the MCTS, the MCTS must be given a sufficient number of simulations to find a suitable solution for training the neural network, if the prior is poor. As shown in \cref{fig:priors_start_node}, the model trained with only 10 MCTS simulations provides no meaningful guidance.
The high number of simulations required, which also increases as the search space grows, means that generating an MCTS training sample takes significantly longer than generating labels for supervised training. The longer running time for MCTS -- combined with the fact that Biggio \emph{et al.} \cite{biggio_neural_2021} show their approach benefits significantly from a large data set with up to 10 million equations -- leads to the conclusion that MCTS is, if at all, only interesting in conjunction  with a pretrained supervised model. A possible use case for MCTS-based approaches could be data with noise, where learning a distribution might be more beneficial than learning a one-hot label.

\textbf{R3} asks whether integrating tabular data sets embeddings can improve the search and how this should be done. \cref{tab:500_generated_equations} clearly shows that integration helps; however, the Text Transformer approach, commonly used in the literature \cite{kamienny_end--end_2022, shojaee_transformer-based_2023, kamienny_deep_2023}, did not perform best, nor did NPT, which performed best in our preliminary experiment with the contrastive loss for tabular data. The best results were achieved with the comparatively simple architectures: CNN, MLP, LSTM, and Bi-LSTM. The reasons for this -- whether the  overfitting of more complex architectures, poor hyperparameters, or other factors -- call for further investigation.

Finally, \textbf{R4} poses the question of whether a grammar should be used to describe the search space or whether it is better to work based on tokens, as many reinforcement learning approaches do. \cref{tab:comparision_grammar_MCTS_} shows that a grammar is effective for using domain knowledge to reduce the search space. In our experiments, we found no evidence that the models have difficulties using the rules from the grammar instead of tokens as output space. On the contrary, when we trained \ours{} with a grammar corresponding to a token-based action space, we either could not learn anything, because the MCTS could not find a solution to train the NN with the given number of simulations, or the number of simulations was so high that we could not sample enough within the training time to train the model. 

\section{Challenges of Neural-Guided Equation Discovery}\label{sec:limitations_and_outview}

In the following, three perspectives -- combinatorial, deep learning, and scientific discovery -- on the problem of equation discovery will be taken and subsequently, the current limitations of \ours{} with regard to these perspectives are analyzed.

The first perspective (\textbf{I}) on equation discovery is to understand it as a combinatorial problem, in the sense of finding an algorithm that is as efficiently as possible with respect to the given resources (time, space, energy, \dots). Its goal is to find the right permutation of tokens for a data set.

The second perspective (\textbf{II}) is to consider equation discovery as a challenging example domain for deep learning. Many other domains (e.g.,  image classification or text translation) in which deep learning is currently successful can be characterized by the fact that the training set and the test set are (more or less) drawn from the same distribution.
One way to achieve better results for these types of domain is to increase the expressive power of the model. This can be done by increasing the number of parameters or using a different architecture. Another way is to use bigger data sets, which effectively reduces the statistical difference between training and test distribution and thereby mitigates the issue of overfitting. 
The approaches in Table \ref{table_approaches} are considering equation discovery as a problem of this type. One problem that must at least be taken into account with these approaches is that equation discovery at the frontier of science  is used for problems where previous experience may not be sufficient.

A third perspective (\textbf{III}) on equation discovery is that of a problem, where new knowledge is generated by using previously gained knowledge. This tradition of consideration goes back to the beginning of AI and tries to answer the question of how we can discover new things with the help of our previous experience or despite it, because the experience is narrowing our view. 

In the following section, we will discuss the limitations of \ours, considering the previously presented perspectives on equation discovery. 
The first limitation of \ours{} in the reinforcement mode is to make it usable as application for other sciences. To do this, the running time for finding an equation must be reduced (\textbf{I}).
A reason for the long running time is that for generating a training sample, a complete MCTS has to be performed. 
To decrease the training time, multiple simulations of MCTS can be run in parallel, and using the NN in inference mode with batch sizes greater than 1 decreases the runtime for the neural-guided MCTS itself. While MCTS and neural guidance can reinforce each other, there is also a risk that they worsen each other (\textbf{I}) (see \cref{sec:experiments:neural_guided_equation_discovery}). If the MCTS alone is not successful, the NN does not receive any meaningful training data and in turn does not provide good guidance in the next iteration of MCTS.
Possible ways to influence the MCTS are reward shaping and changing the hyperparameters used in the MCTS (\textbf{I}). 
Other approaches could be curriculum learning and reducing the interdependency of the two components by using more supervised learning (\textbf{II}). In the field of neural networks, an open question remains how the information in the data tables can be made reliably accessible to a neural net (\textbf{II}). For the already proposed architectures, the question of scaling to a higher number of variables is an open problem (\textbf{II}), while taking into account  invariances with respect to permutations of rows and columns.  
Finally, from the third area, the question arises whether we can learn not only to fit individual data sets, but also -- in the sense of meta-learning -- learn how to fit data sets (\textbf{III})? Can known concepts, such as symmetry, integration, or energy conservation, automatically be derived and so far unknown concepts be discovered (\textbf{III})? 

\section{Conclusion}\label{sec:Conclusion}

We introduce MGMT, a modular equation discovery system with a neural-guided MCTS that can be trained using supervised training or reinforcement learning. Comparing both approaches for various architectures to embed tabular data, we found that supervised learning outperforms reinforcement learning in nearly all cases. Using the rules of a context-free grammar instead of tokens as action space is an easy but effective way to introduce domain knowledge into the search, as has been shown already a long time ago for other equation discovery schemes  %
\cite{TodorovskiDzeroski1997}.
Regarding the embedding of tabular data, we found that NPT, a permutation-resistant transformer architecture, delivers the best results for the auxiliary task of contrastive learning for tabular data sets. However, when testing \ours{}, simpler architectures like LSTMs showed better results.

In future work, we want to explore which patterns are currently used when embedding tabular data sets, how embedding architectures can capture the underlying mathematical functions, and how these can be applied to high-di\-men\-sio\-nal tables. Additionally, it will be interesting to see whether fine-tuning supervised-trained models with reinforcement learning for noisy data brings an advantage.

\section*{Acknowledgements}
This research project was partly funded by the Hessian Ministry of Higher Education, Research, Science and the Arts (HMWK) within the projects \textit{The Third Wave of Artificial Intelligence} (3AI) and \textit{hessian.AI}.

\bibliographystyle{unsrt}%
\bibliography{library.bib}
\newpage
\appendix
\section{Appendix}

\begin{multicols}{2}
\small
\begin{align*} 
\noalign{\textbf{Grammar A}}\\
1.\!\!\!\!\!\!\!\!\!\!\!\!&&N     &=          &&\!\!\!\!\{\mathit{S}, \mathit{Variable}, \mathit{Power}\}\\ 
2.\!\!\!\!\!\!\!\!\!\!\!\!&&T     &=          &&\!\!\!\!\{+, -, \cdot, /, \sin, \cos, \log, ^\wedge{}, x_0, x_1, c\}\\
3.\!\!\!\!\!\!\!\!\!\!\!\!&&R     &=          &&\!\!\!\!\{ \\
    &&[0.05] &\phantom{=}&&\!\!\!\!\!\!\!\!\mathit{S}\rightarrow{} + \, \mathit{c} \, \mathit{Variable}\\
    &&[0.05]&\phantom{=}&&\!\!\!\!\!\!\!\!\mathit{S}\rightarrow{} + \, \mathit{c} \, ^\wedge{} \, \mathit{Power} \, \mathit{Variable}\\
    &&[0.05] &\phantom{=}&&\!\!\!\!\!\!\!\!\mathit{S}\rightarrow{} + \, \mathit{c} \, \sin \, \mathit{Variable} \\
    &&[0.05] &\phantom{=}&&\!\!\!\!\!\!\!\!\mathit{S}\rightarrow{} + \, \mathit{c} \, \cos \, \mathit{Variable} \\
    &&[0.05]&\phantom{=}&&\!\!\!\!\!\!\!\!\mathit{S}\rightarrow{} + \, \mathit{c} \, ^\wedge{} \, \mathit{Power} \, \mathit{Variable}\\
    &&[0.05] &\phantom{=}&&\!\!\!\!\!\!\!\!\mathit{S}\rightarrow{} - \, \mathit{c} \, \cdot \, \mathit{c} \, / \, 1 \, + \, ^\wedge{} \, 2 \, \mathit{Variable} \, 1\\
    &&[0.05] &\phantom{=}&&\!\!\!\!\!\!\!\!\mathit{S}\rightarrow{} / \,  \mathit{c} \,  \mathit{Variable}  \\
    &&[0.05] &\phantom{=}&&\!\!\!\!\!\!\!\!\mathit{S}\rightarrow{} / \, \mathit{c} \,^\wedge{} \, \mathit{Variable} \, \mathit{c}  \\
    &&[0.05] &\phantom{=}&&\!\!\!\!\!\!\!\!\mathit{S}\rightarrow{} + \, \mathit{c} \, \mathit{ln} \, \mathit{Variable}  \\
    &&[0.05] &\phantom{=}&&\!\!\!\!\!\!\!\!\mathit{S}\rightarrow{} ^\wedge{} \, 0.5 \, \cdot \, \mathit{c} \, ^\wedge{} \, \mathit{Power} \, \mathit{Variable} \\
    &&[0.05] &\phantom{=}&&\!\!\!\!\!\!\!\!\mathit{S}\rightarrow{} ^\wedge{} \, ^\wedge{} \, 3 \, \mathit{Variable} \,  \mathit{c} \\
    &&[0.04] &\phantom{=}&&\!\!\!\!\!\!\!\!\mathit{S}\rightarrow{} + \, \mathit{c} \, ^\wedge{} \, - \, 0 \, ^\wedge{} \, \mathit{Power} \, \mathit{Variable} \, 2\\
    &&[0.04] &\phantom{=}&&\!\!\!\!\!\!\!\!\mathit{S}\rightarrow{} / \, 1 \, + \, 1 \, ^\wedge{} \, \mathit{Variable} \, \mathit{c}\\
    &&[0.04] &\phantom{=}&&\!\!\!\!\!\!\!\!\mathit{S}\rightarrow{} + \, \mathit{c} \, ^\wedge{} \, \mathit{Power} \, \mathit{Variable} \\
    &&[0.04] &\phantom{=}&&\!\!\!\!\!\!\!\!\mathit{S}\rightarrow{} - \, 1 \, + \, \cdot \, \mathit{c} \, ^\wedge{} \, 3 \, \mathit{Variable} \, + \, \cdot \, \mathit{c} \, ^\wedge{} \, 2 \, \mathit{Variable} \,  \cdot \, \mathit{c} \, \mathit{Variable} \\
    &&[0.04] &\phantom{=}&&\!\!\!\!\!\!\!\!\mathit{S}\rightarrow{} + \, \mathit{c} \, \sin \, \cdot \,  2 \, \mathit{Variable}  \\
    &&[0.05] &\phantom{=}&&\!\!\!\!\!\!\!\!\mathit{S}\rightarrow{} + \, \mathit{c} \, \cos \, \cdot \, 2 \, \mathit{Variable}  \\
    &&[0.05] &\phantom{=}&&\!\!\!\!\!\!\!\!\mathit{S}\rightarrow{} + \cdot \, \mathit{c} \, ^\wedge{} \, \mathit{Power} \, \mathit{Variable} \,  + \, \cdot \, \mathit{c} \, ^\wedge{} \, \mathit{Power} \, \mathit{Variable}  \,  \cdot \, \mathit{c} \, \mathit{Variable} \\
    &&[0.05] &\phantom{=}&&\!\!\!\!\!\!\!\!\mathit{S}\rightarrow{} + \cdot \, \mathit{c} \, ^\wedge{} \, \mathit{Power} \, \mathit{Variable} \,  + \, \cdot \, \mathit{c} \, ^\wedge{} \,  \mathit{Power} \, \mathit{Variable} \,  + \, \cdot \, \mathit{c}  \, ^\wedge{} \, \mathit{Power} \, \mathit{Variable} \,   \cdot \, \mathit{c} \, \mathit{Variable}   \\
    &&[0.05] &\phantom{=}&&\!\!\!\!\!\!\!\!\mathit{S}\rightarrow{}  - \, \mathit{c} \, \mathit{Variable} \\
    &&[0.05] &\phantom{=}&&\!\!\!\!\!\!\!\!\mathit{S}\rightarrow{}  - \, \mathit{c} \, ^\wedge{} \, \mathit{Power} \, \mathit{Variable} \\
    &&[0.2] &\phantom{=}&&\!\!\!\!\!\!\!\!\mathit{Power}\rightarrow{} 0.33 \, | \, 0.5 \, | \, 2 \, | \,  3 \, | \, 4  \\
    &&[0.5] &\phantom{=}&&\!\!\!\!\!\!\!\!\mathit{Variable}\rightarrow{}  x_0 \,| \, x_1 \\
    && \phantom{A}     &\phantom{=}&&\!\!\!\!\}\\
4.\!\!\!\!\!\!\!\!\!\!\!\!&&S     &=          &&\!\!\!\!\!\mathit{Start}.
\end{align*}

\columnbreak

\begin{align*} 
\noalign{ }\\
\end{align*}

\end{multicols}

\newpage
\begin{multicols}{2}
\small

\begin{align*} 
\noalign{\textbf{Grammar B}}\\
1.\!\!\!\!\!\!\!\!\!\!\!\!&&N     &=          &&\!\!\!\!\{\mathit{S}, \mathit{Inner}, \mathit{I}\}\\ 
2.\!\!\!\!\!\!\!\!\!\!\!\!&&T     &=          &&\!\!\!\!\{+, -, \cdot, \sin, \cos, \log, ^\wedge{}, x_0, x_1, c\}\\
3.\!\!\!\!\!\!\!\!\!\!\!\!&&R     &=          &&\!\!\!\!\{ \\
    &&[0.15] &\phantom{=}&&\!\!\!\!\!\!\!\!\mathit{S}\rightarrow{} + \mathit{S} \, \mathit{S}\\
    &&[0.15]&\phantom{=}&&\!\!\!\!\!\!\!\!\mathit{S}\rightarrow{} - \mathit{S} \, \mathit{S}\\
    &&[0.1] &\phantom{=}&&\!\!\!\!\!\!\!\!\mathit{S}\rightarrow{} \cdot \mathit{S} \, \mathit{S}\\
    &&[0.02]&\phantom{=}&&\!\!\!\!\!\!\!\!\mathit{S}\rightarrow{} ^\wedge{} 6 x_0\\
    &&[0.03]&\phantom{=}&&\!\!\!\!\!\!\!\!\mathit{S}\rightarrow{} ^\wedge{} 5 x_0\\
    &&[0.03]&\phantom{=}&&\!\!\!\!\!\!\!\!\mathit{S}\rightarrow{} ^\wedge{} 4 x_0\\
    &&[0.03]&\phantom{=}&&\!\!\!\!\!\!\!\!\mathit{S}\rightarrow{} ^\wedge{} 3 x_0\\
    &&[0.05]&\phantom{=}&&\!\!\!\!\!\!\!\!\mathit{S}\rightarrow{} ^\wedge{} 2 x_0\\
    &&[0.02]&\phantom{=}&&\!\!\!\!\!\!\!\!\mathit{S}\rightarrow{} ^\wedge{} x_1 x_0\\
    &&[0.1] &\phantom{=}&&\!\!\!\!\!\!\!\!\mathit{S}\rightarrow{} x_0\\
    &&[0.1] &\phantom{=}&&\!\!\!\!\!\!\!\!\mathit{S}\rightarrow{} x_1\\
    &&[0.1]&\phantom{=}&&\!\!\!\!\!\!\!\!\mathit{S}\rightarrow{} c\\
    &&[0.03]&\phantom{=}&&\!\!\!\!\!\!\!\!\mathit{S}\rightarrow{} \sin \mathit{ Inner}\\
    &&[0.03]&\phantom{=}&&\!\!\!\!\!\!\!\!\mathit{S}\rightarrow{} \cos \mathit{ Inner}\\
    &&[0.03]&\phantom{=}&&\!\!\!\!\!\!\!\!\mathit{S}\rightarrow{} \log \mathit{ Inner}\\
    &&[0.3] &\phantom{=}&&\!\!\!\!\!\!\!\!\mathit{Inner}\rightarrow{} + \mathit{I} \, \mathit{I}\\
    &&[0.3] &\phantom{=}&&\!\!\!\!\!\!\!\!\mathit{Inner}\rightarrow{} \cdot \mathit{I} \, \mathit{I}\\
    &&[0.4] &\phantom{=}&&\!\!\!\!\!\!\!\!\mathit{Inner}\rightarrow{} \mathit{I}\\
    &&[0.2] &\phantom{=}&&\!\!\!\!\!\!\!\!\mathit{I}\rightarrow{} x_0\\
    &&[0.2] &\phantom{=}&&\!\!\!\!\!\!\!\!\mathit{I}\rightarrow{} x_1\\
    &&[0.2] &\phantom{=}&&\!\!\!\!\!\!\!\!\mathit{I}\rightarrow{} c\\
    &&[0.2] &\phantom{=}&&\!\!\!\!\!\!\!\!\mathit{I}\rightarrow{} ^\wedge{} 2 x_0\\
    &&[0.2] &\phantom{=}&&\!\!\!\!\!\!\!\!\mathit{I}\rightarrow{} ^\wedge{} 2 x_1\\    
    && \phantom{A}     &\phantom{=}&&\!\!\!\!\}\\
4.\!\!\!\!\!\!\!\!\!\!\!\!&&S     &=          &&\!\!\!\!\!\mathit{Start}.
\end{align*}

\columnbreak

\begin{align*} 
\noalign{\textbf{Grammar C}}\\
1.\!\!\!\!\!\!\!\!\!\!\!\!\!\!&&N     &=          &&\!\!\!\!\{\mathit{S}, \mathit{I}\}\\ 
2.\!\!\!\!\!\!\!\!\!\!\!\!\!\!&&T     &=          &&\!\!\!\!\{+, -, \cdot, \sin, \cos, \log, ^\wedge{}, x_0, x_1, c\}\\
3.\!\!\!\!\!\!\!\!\!\!\!\!\!\!&&R     &=          &&\!\!\!\!\{ \\
    &&[0.15] &\phantom{=}&&\!\!\!\!\!\!\!\!\mathit{S}\rightarrow{} + \mathit{S} \, \mathit{S}\\
    &&[0.1] &\phantom{=}&&\!\!\!\!\!\!\!\!\mathit{S}\rightarrow{} - \mathit{S} \, \mathit{S}\\
    &&[0.15] &\phantom{=}&&\!\!\!\!\!\!\!\!\mathit{S}\rightarrow{} \cdot \mathit{S} \, \mathit{S}\\
    &&[0.1] &\phantom{=}&&\!\!\!\!\!\!\!\!\mathit{S}\rightarrow{} / \mathit{S} \,  \mathit{S}\\
    &&[0.025] &\phantom{=}&&\!\!\!\!\!\!\!\!\mathit{S}\rightarrow{} ^\wedge{}\mathit{S} \, \mathit{S}\\
    &&[0.05] &\phantom{=}&&\!\!\!\!\!\!\!\!\mathit{S}\rightarrow{} ^\wedge{} 2 x_0\\
    &&[0.05] &\phantom{=}&&\!\!\!\!\!\!\!\!\mathit{S}\rightarrow{} ^\wedge{} 2 x_1\\
    &&[0.1] &\phantom{=}&&\!\!\!\!\!\!\!\!\mathit{S}\rightarrow{} x_0\\
    &&[0.1] &\phantom{=}&&\!\!\!\!\!\!\!\!\mathit{S}\rightarrow{} x_1\\
    &&[0.1] &\phantom{=}&&\!\!\!\!\!\!\!\!\mathit{S}\rightarrow{} c\\
    &&[0.025] &\phantom{=}&&\!\!\!\!\!\!\!\!\mathit{S}\rightarrow{} \sin\mathit{ I}\\
    &&[0.025] &\phantom{=}&&\!\!\!\!\!\!\!\!\mathit{S}\rightarrow{} \cos\mathit{ I}\\
    &&[0.025] &\phantom{=}&&\!\!\!\!\!\!\!\!\mathit{S}\rightarrow{} \log \mathit{ I}\\
    &&[0.2] &\phantom{=}&&\!\!\!\!\!\!\!\!\mathit{I}\rightarrow{} + \mathit{I} \, \mathit{I}\\
    &&[0.2] &\phantom{=}&&\!\!\!\!\!\!\!\!\mathit{I}\rightarrow{} \cdot \mathit{I} \, \mathit{I}\\
    &&[0.1] &\phantom{=}&&\!\!\!\!\!\!\!\!\mathit{I}\rightarrow{} / \mathit{I} \, \mathit{I}\\
    &&[0.2] &\phantom{=}&&\!\!\!\!\!\!\!\!\mathit{I}\rightarrow{} x_0\\
    &&[0.2] &\phantom{=}&&\!\!\!\!\!\!\!\!\mathit{I}\rightarrow{} x_1\\
    &&[0.1] &\phantom{=}&&\!\!\!\!\!\!\!\!\mathit{I}\rightarrow{} c\\
    && \phantom{A}     &\phantom{=}&&\!\!\!\!\}\\
4.\!\!\!\!\!\!\!\!\!\!\!\!\!\!&&S     &=          &&\!\!\!\!\!\mathit{Start}.
\end{align*}

\end{multicols}

{
 \centering
\begin{table}[tp]
   
\caption{\textbf{Mean number of simulations until an equation with $r > 0.999$ is found for each equation in the Nguyen dataset}. The mean ($\mu$) and standard deviation ($\sigma$) are calculated with 5 seeds for Grammar B, C, and  DSO. AmEx-MCTS can find the equation with fewer simulations in most cases. 
If the dataset used in DSO \cite{Deep_symbolic_regression_Recovering_mathematical_expressions_from_data_via_risk_seeking_policy_gradients} did not specify the second variable, it was filled with 0.
}
\begin{tabularx}{\textwidth}{Xrrrrrr}
{} &                                 \multicolumn{3}{c}{Equation} & \multicolumn{3}{c}{DSO} \\
{} & &&&      \multicolumn{1}{c}{$\mu$} &  \multicolumn{1}{c}{$\sigma$} &  Failed runs\\
\midrule
Nguyen 1   & \multicolumn{3}{c}{ $x_0^3 + x_0^2 + x_0 $}  &   92400 &    16103 &        0 \\
Nguyen 2   & \multicolumn{3}{c}{$x_0^4 + x_0^3 + x_0^2 + x_0 $}  &  161400 &    26063 &        0 \\
Nguyen 3   & \multicolumn{3}{c}{$x_0^5 + x_0^4 +  x_0^3 + x_0^2 + x_0 $ } &  179000 &    32481 &        0 \\
Nguyen 4   & \multicolumn{3}{c}{$x_0^6 + x_0^5 + x_0^4 + x_0^3 + x_0^2 + x_0 $}  &  169800 &    40481 &        0 \\
Nguyen 5   & \multicolumn{3}{c}{$\sin (x_0^2) + \cos(x_0) -1  $ } &  473000 &   162635 &        3 \\
Nguyen 6   &  \multicolumn{3}{c}{$\sin (x_0) + \sin (x_0 + x_0^2)$ } &  102600 &    23448 &        0 \\
Nguyen 7   &  \multicolumn{3}{c}{$\log (x_0 + 1) + \log (x_0^2 + 1)$}  &       - &        - &        5 \\
Nguyen 8   & \multicolumn{3}{c}{$\sqrt{x_0}$}  &  194200 &    68584 &        0 \\
Nguyen 9   &  \multicolumn{3}{c}{ $\sin(x_0) + sin(x_1^2)$}  &  149800 &    49277 &        0 \\
Nguyen 10  & \multicolumn{3}{c}{$2 \cdot sin (x_0) \cdot cos(x_1)$}  &  355000 &    96449 &        0 \\
Nguyen 11  & \multicolumn{3}{c}{ $x_0^{x_1}$}  &   48500 &    31932 &        1 \\
Nguyen 12  & \multicolumn{3}{c}{ $x_0^4 - x_0^3 + 0.5 \cdot x_1^2 - x_1$}  &       - &        - &        5 \\
\bottomrule

{} & \multicolumn{3}{l}{Classic MCTS + Grammar B} & \multicolumn{3}{l}{AmEx-MCTS + Grammar B } \\
{} &                     \multicolumn{1}{c}{$\mu$} &  \multicolumn{1}{c}{$\sigma$} &  Failed runs&                   \multicolumn{1}{c}{$\mu$} &  \multicolumn{1}{c}{$\sigma$} &  Failed runs\\
\midrule
Nguyen 1   &                   4971 &     4015 &        0 &                 \textbf{1851} &     1153 &        0 \\
Nguyen 2   &                  17578 &    18036 &        0 &                 \textbf{7956} &     6107 &        0 \\
Nguyen 3   &                  57988 &    49268 &        3 &                \textbf{46860} &    12985 &        1 \\
Nguyen 4   &                  \textbf{45751} &    33060 &        0 &                55686 &    29716 &        2 \\
Nguyen 5   &                   6005 &     4979 &        0 &                 \textbf{1605} &      813 &        0 \\
Nguyen 6   &                  20122 &    15791 &        0 &                \textbf{15808} &    11535 &        0 \\
Nguyen 7   &                   2600 &     2410 &        0 &                 \textbf{1171} &      477 &        0 \\
Nguyen 8   &                      \textbf{9} &        2 &        0 &                    \textbf{8} &        4 &        0 \\
Nguyen 9   &                   \textbf{4454} &     4430 &        0 &                 7400 &     3315 &        0 \\
Nguyen 10  &                  37258 &    24607 &        0 &                \textbf{27407} &    12185 &        0 \\
Nguyen 11  &                      \textbf{5} &        4 &        0 &                    \textbf{5} &        3 &        0 \\
Nguyen 12  &                  44946 &    31519 &        0 &                \textbf{14897} &    16392 &        0 \\
\bottomrule

{} & \multicolumn{3}{l}{Classic MCTS + Grammar C} & \multicolumn{3}{l}{AmEx-MCTS + Grammar C } \\
{} &                     \multicolumn{1}{c}{$\mu$} &  \multicolumn{1}{c}{$\sigma$} &  Failed runs&                   \multicolumn{1}{c}{$\mu$} &  \multicolumn{1}{c}{$\sigma$} &  Failed runs\\
\midrule
Nguyen 1   &                  17745 &    16186 &        0 &                \textbf{15190} &    20227 &        0 \\
Nguyen 2   &                  21432 &    17290 &        0 &                \textbf{19995} &    21928 &        0 \\
Nguyen 3   &                  \textbf{20990} &    12140 &        1 &                37071 &    21654 &        2 \\
Nguyen 4   &                  14060 &     9212 &        1 &                \textbf{11230} &    12179 &        0 \\
Nguyen 5   &                   2799 &     1884 &        0 &                 \textbf{1842} &     1904 &        0 \\
Nguyen 6   &                  32958 &    16118 &        3 &                \textbf{23670} &    24879 &        1 \\
Nguyen 7   &                   1153 &      777 &        0 &                  \textbf{477} &      323 &        0 \\
Nguyen 8   &                    316 &      279 &        0 &                  \textbf{199} &      189 &        0 \\
Nguyen 9   &                  10114 &     6309 &        0 &                 \textbf{8612} &     6305 &        0 \\
Nguyen 10  &                  19026 &    11318 &        1 &                \textbf{14570} &     6137 &        0 \\
Nguyen 11  &                    595 &      630 &        0 &                  \textbf{490} &      484 &        0 \\
Nguyen 12  &                  29595 &    29813 &        0 &                \textbf{18530} &    19563 &        1 \\
\bottomrule
\end{tabularx}

    \label{tab:detail_nguyen_result}
\end{table}
}

\newcolumntype{Y}{>{\centering\arraybackslash}X}
\newcolumntype{L}{>{\RaggedRight\arraybackslash}X}

\begin{table}[ht]
\caption{\label{tab:hyperparameter}\textbf{Most important hyperparameter for the experiments}. All hyperparameters are given in the repository https://anonymous.4open.science/r/EquationFinder-7C3 .}
\begin{tabularx}{\textwidth}{LYYY}
                        &   \makecell[c]{6.1 \\ Data Set \\ Embeddings \\ \ } &  \makecell[c]{ 6.2\\ Neural-Guided \\ Equation \\ Discovery}  &\makecell[c]{ 6.3\\ AmEx-MCTS \\ vs. \\ Classic MCTS} \\
\midrule
Grammar                              & A                                                                     & A                                                                      & B and C                                     \\
\hline
Seed                                      & 1 - 5                                                                 & 1                                                                      & 0 - 4                                         \\
\hline
Iterations                                & 300                                                                   & 200                                                                    & -                                           \\
\hline
Training Mode                              & Contrastive Loss                          & Supervised, MCTS 250, MCTS 125, MCTS 50, MCTS 10, Uniform Distribution & Uniform Distribution                        \\
\hline
Equations per Iteration                   & 50                                                                    & 50                                                                     & 1                                           \\
\hline
Equations in Test Set               & 10 every 10 iterations                                                & 500                                                                    & Nguyen benchmark                            \\
\hline
Minimum Reward                            & -1                                                                    & -1                                                                     & -1                                          \\
\hline
Maximum Depth of Tree                     & 10                                                                    & 10                                                                     & 10                                          \\
\hline
Maximum Number of Nodes                 & 25                                                                    & 25                                                                     & 25                                          \\
\hline
Maximum Number of Constants               & 5                                                                     & 5                                                                      & 2                                           \\
\hline
Batch Size Training                       & 64                                                                    & 64                                                                     & -                                           \\
\hline
Architecture for Table Embedding          &\multicolumn{2}{c}{\makecell[c]{Bi-LSTM, LSTM,  CNN,  MLP, \\NPT, Text Transformer, No Embedding }}       & -                                           \\
\hline
With Syntax Tree Embedding                & True(No effect)                                                       & True, False                                                            & -                                           \\
\hline
Rows per Data set                          & 100                                                                   & 100                                                                    & 20                                          \\
\hline
c in PUCT                                 & -                                                                    & 10                                                                     & 10                                          \\
\hline
$\lambda$ in Contrastive Loss & 0.1         
& -                                                                      & -          \\                                 
\bottomrule
\end{tabularx}
\end{table}

\end{document}